\documentclass{article}
\usepackage{amsthm}

\usepackage{arxiv}
\theoremstyle{definition}

\usepackage[utf8]{inputenc} % allow utf-8 input
\usepackage[T1]{fontenc}    % use 8-bit T1 fonts
\usepackage{hyperref}       % hyperlinks
\usepackage{url}
\usepackage{booktabs}
\usepackage{multirow}
\usepackage{xcolor}
\usepackage{colortbl}
\usepackage{array}
\usepackage{amssymb}
\usepackage{comment}
\usepackage{latexsym}
\usepackage{amsmath}
\usepackage{enumitem}
\usepackage{cite}
\usepackage{multirow}
\usepackage{bm}
\usepackage{makecell}
\usepackage{soul}
\usepackage[ruled,vlined,linesnumbered]{algorithm2e}
\usepackage{booktabs}       % professional-quality tables
\usepackage{amsfonts}       % blackboard math symbols
\usepackage{nicefrac}       % compact symbols for 1/2, etc.
\usepackage{microtype}      % microtypography
\usepackage{lipsum}
\usepackage{graphicx}
\usepackage{subcaption}
\usepackage{xcolor}
\usepackage{algpseudocode}

\usepackage{graphicx}
\usepackage{subcaption}
\usepackage[percent]{overpic}
\usepackage{caption}
\graphicspath{ {./images/} }

\newcommand{\be}{\begin{eqnarray}}
\newcommand{\ee}{\end{eqnarray}}

\usepackage{amsfonts} % For \mathbb{C}
\usepackage{tablefootnote} % For table-specific footnotes

\definecolor{codeblue}{RGB}{0, 82, 147}
\definecolor{codegreen}{RGB}{0, 128, 0}
\definecolor{codegray}{RGB}{100, 100, 100}
\definecolor{codeorange}{RGB}{230, 145, 56}
\definecolor{darkerblue}{rgb}{0,0.08,0.45}
\definecolor{royalblue}{RGB}{65,105,225}
\definecolor{lightblue}{RGB}{221,235,247}
\definecolor{fig3blue}{RGB}{47, 122, 232}
\definecolor{fig3red}{RGB}{213, 32, 52}
\definecolor{fig3green}{RGB}{0, 137, 72}
\definecolor{fig3yellow}{RGB}{217, 161, 5}
\definecolor{gray94}{gray}{.94}
\definecolor{gray90}{gray}{.90}
\definecolor{darkgreen}{RGB}{34,139,34}

\title{From Non-Convex Self-Concordant Regularization to Scalable Quasi-Newton Training of PINNs}

\author{
  Chenhao Si \\
  School of Data Science \\
  The Chinese University of Hong Kong, Shenzhen \\
  Shenzhen, China \\
  \texttt{chenhao.si@link.cuhk.edu.cn}
  \And
  Kang An \\
  Department of Computational Applied Mathematics \\
  and Operations Research \\
  Rice University \\
  Houston, USA \\
  \texttt{kang.an@rice.edu}
  \And
  Shiqian Ma \\
  Department of Computational Applied Mathematics \\
  and Operations Research \\
  Rice University \\
  Houston, USA \\
  \texttt{shiqian.ma@rice.edu}
  \And
  Ming Yan\footnotemark[1] \\
  School of Data Science \\
  The Chinese University of Hong Kong, Shenzhen \\
  Shenzhen, China \\
  \texttt{yanming@cuhk.edu.cn}
}

\begin{document}
\maketitle
\footnotetext[1]{Corresponding author.}
\begin{abstract}
Physics-informed neural networks (PINNs) often require high-accuracy quasi-Newton refinement to obtain reliable partial differential equation solutions, but their residual objectives can exhibit indefinite, nearly singular, and poorly scaled local curvature. Regularized quasi-Newton methods provide established mechanisms for stabilizing secant models, while self-concordant methods provide local-metric rules for curvature-dependent step selection. Building on these two lines of work, we propose \textsc{SCORE}, a self-concordance-inspired quasi-Newton method with decrement-coupled shifted secant geometry for PINN training. Its distinguishing mechanism is that a single quasi-Newton decrement computed from the learned inverse metric jointly determines a strong-Wolfe-tested candidate step and an adaptive shift used to define the next secant geometry. The shifted displacement represents the action of an averaged shifted metric along the accepted step, while requiring neither Hessian construction nor Hessian-vector products. Under a local spectral-equivalence condition, we show that the quasi-Newton decrement and candidate step remain comparable to their counterparts in a positive shifted metric, and recover the normalized self-concordant rule in the matched-metric case. Strong Wolfe acceptance, fallback line search, and standard curvature safeguards provide globalization without modifying the underlying PINN objective. Experiments on the viscous Burgers, Kuramoto--Sivashinsky, Korteweg--de Vries, and complex Ginzburg--Landau equations show that \textsc{SCORE} attains lower final errors than the tested BFGS and self-scaled Broyden baselines. The Burgers ablation further indicates that shifted curvature stabilization and decrement-based step selection make complementary contributions to high-accuracy refinement.
\end{abstract}

\section{Introduction}
\label{sec:introduction}

Physics-informed neural networks (PINNs)~\cite{raissi2019physics,karniadakis2021physics} have become a widely used framework for scientific machine learning problems governed by partial differential equations (PDEs), alongside operator-learning approaches with improved expressivity for PDE solution maps~\cite{gao2025dynamic}. By incorporating PDE residuals, boundary conditions, and initial conditions into the training objective, PINNs provide a mesh-free formulation for both forward simulation and inverse parameter identification, often requiring limited or no observational data; related neural-operator studies have also examined discretization mismatch across resolutions~\cite{gao2025discretization}. They have been applied to heat transfer~\cite{xu2023physics,cai2021physics,si2025initialization,majumdar2025hxpinn}, solid mechanics~\cite{hu2024physics,faroughi2024physics}, magnetic anomaly detection~\cite{besnard2025fast}, petroleum transport modeling~\cite{wang2023investigation,wang2025investigation,wang2026progress}, stochastic systems~\cite{zhang2020learning,chen2021solving}, and uncertainty quantification~\cite{yang2019adversarial,gao2023active, zhang2019quantifying,yang2021b}. More broadly, recent machine-learning studies have emphasized that predictive accuracy alone may be insufficient when learned models must respect application-specific structure or interpretability requirements~\cite{wang2026embeddingfoundationmodelpredictions,wang2026auditingfixingeconomicvalidity}. Related reliability issues also arise under class imbalance~\cite{han2026validationstagecombinatorialfusionanalysis}, high-cardinality representations~\cite{han2026interpretablevslearnedencoders}, incomplete signals, and changing observation windows~\cite{han2026earlyearlyenoughdesigndependent,wang2026timeseriesfoundationmodel,yuan2022opticalflow}. These considerations are particularly relevant to scientific machine learning, where predictive accuracy must coexist with governing physical constraints, including symmetry-aware representations~\cite{gao2024coordinate} and structure-preserving dynamics learning~\cite{xu2025velocity}. Despite their broad applicability, training PINNs to high accuracy remains difficult, particularly when reliable PDE solutions require the residual errors to be reduced to very small levels.

PINN training can be affected by stiffness, imbalance among loss terms, gradient pathologies, spectral bias, and sensitivity to the optimizer and its parameters~\cite{wang2021understanding,si2026convolution,krishnapriyan2021characterizing,rathore2024challenges,an2026lightweight,si2026complex}. First-order methods such as Adam~\cite{kingma2014adam} are widely used during the early stage of training because of their robustness and low per-iteration cost. However, they may converge slowly or stagnate when the residual objective becomes severely ill-conditioned. This behavior has motivated the use of second-order and curvature-based optimizers for PINNs, including quasi-Newton methods such as L-BFGS, BFGS, self-scaled BFGS, and self-scaled Broyden variants~\cite{rathore2024challenges,urban2025unveiling,jnini2026curvature,kiyani2025optimizing}, as well as Newton-type, Gauss--Newton, Levenberg--Marquardt, natural-gradient, and structured preconditioned methods~\cite{jnini2025gauss,muller2023achieving,jnini2025dual,levenberg1944method,dai2026tinns,alizadeh2025physics,shahab2026physics}. These studies suggest that local curvature information can improve both the optimization efficiency and the attainable accuracy of PINNs.

Using curvature information reliably is nevertheless challenging. Although a PINN objective is typically written as a sum of squared residuals, it is generally nonconvex. Its local curvature contains a nonnegative Gauss--Newton contribution together with an additional residual-dependent contribution that may have either sign. The resulting curvature model can therefore be indefinite, nearly singular, or poorly conditioned. These difficulties are particularly relevant to quasi-Newton methods, which infer curvature from successive parameter steps and gradient differences rather than forming the Hessian explicitly. When the observed secant curvature is weak or poorly scaled, the learned approximation can become unreliable, and the optimizer may require substantial damping, update skipping, or conservative line-search steps~\cite{rathore2024challenges,urban2025unveiling,jnini2026curvature,kiyani2025optimizing}.

These safeguards are essential for robust quasi-Newton refinement, but the residual least-squares structure of PINNs raises a more fundamental question: what local curvature geometry should a self-scaled quasi-Newton method learn when the full residual Hessian is indefinite, nearly singular, or poorly conditioned? Self-scaling adapts the magnitude of the inverse approximation to changes in the observed curvature. The secant update, however, is still driven by the raw gradient displacement and therefore reflects both the nonnegative Gauss--Newton contribution and the sign-indefinite residual-dependent contribution of the PINN Hessian.

Self-concordance provides a curvature-relative perspective on this problem. Classical self-concordance measures curvature variation using the local Hessian metric and uses the corresponding decrement to calibrate the Newton step~\cite{nesterov1994interior,nesterov2013introductory,bach2010self,sun2019generalized}. For a nonconvex PINN objective, a positive local metric is obtained through the shifted curvature geometry of weak self-concordance~\cite{goldfarb2025non}. The shift supplies a positive curvature floor, while the associated decrement measures the proposed step in the resulting local geometry. Regularization therefore emerges naturally as the mechanism that realizes curvature-relative control for a nonconvex objective.

Regularized quasi-Newton methods provide Hessian-free mechanisms for representing positively shifted curvature models through modified secant information~\cite{tankaria2022regularized,kanzow2023regularization}. Self-concordant quasi-Newton methods further demonstrate how local curvature information can guide step selection~\cite{gao2019quasi}. These developments provide the optimization foundations for incorporating a positive local metric and curvature-dependent step control into secant-based methods.

In this work, we use these principles to adapt self-scaled Broyden refinement to the residual geometry of PINNs. We propose \textsc{SCORE}, a self-concordance-inspired SSBroyden method in which the shifted local metric is represented through the secant displacement used by the inverse update. A quasi-Newton decrement computed from the learned inverse metric determines both an adaptive curvature shift and a self-concordance-inspired candidate step. The candidate is tested by the strong Wolfe conditions, with a standard Wolfe line search used whenever the candidate is rejected.

The resulting construction connects three aspects of high-accuracy PINN optimization within a single local geometry. The PINN residual structure identifies the source of indefinite and poorly scaled curvature; weak self-concordance motivates a positive shifted metric and decrement-based control; and SSBroyden provides a scalable secant representation of this metric. The same learned-metric decrement coordinates the step taken with the current curvature model and the stabilization used to construct the next one.

Our main contributions are summarized as follows:
\begin{itemize}
    \item We relate the residual least-squares geometry of PINNs to the curvature information learned during self-scaled Broyden refinement. The interaction between the nonnegative Gauss--Newton component and the residual-dependent Hessian component motivates stabilizing the local metric represented by the secant update.

    \item We introduce a weak-self-concordance-based local-metric perspective for high-accuracy PINN refinement. The resulting positive shifted geometry provides a unified basis for curvature stabilization and decrement-dependent step control in a nonconvex residual objective.

    \item We develop \textsc{SCORE}, a Hessian-free realization of this perspective within SSBroyden. The shifted secant displacement represents the positive local metric along the accepted step, while a quasi-Newton decrement jointly determines the adaptive curvature shift and a strong-Wolfe-tested candidate step.

    \item We evaluate \textsc{SCORE} on the viscous Burgers, Kuramoto--Sivashinsky, Korteweg--de Vries, and complex Ginzburg--Landau equations. The method consistently attains lower final errors than the tested BFGS and SSBroyden baselines, and the Burgers ablation demonstrates complementary benefits from shifted curvature stabilization and decrement-based step selection.
\end{itemize}

\section{Preliminaries and Related Work}
\label{sec:preliminaries}

This section introduces the three ingredients used in the development of \textsc{SCORE}. We first formulate PINN training as a residual least-squares problem and describe its local curvature. We then review secant-based quasi-Newton methods, which approximate curvature without explicitly forming the Hessian. Finally, we summarize weak self-concordance and the shifted curvature metric used in regularized Newton methods. Section~\ref{sec:motivation} combines these ideas to motivate our shifted quasi-Newton construction.

\subsection{PINN Objectives as Residual Least-Squares Problems}
\label{Sec PINN}

Physics-informed neural networks enforce differential equations through residual terms evaluated at collocation points~\cite{raissi2019physics,karniadakis2021physics}. Let \(\Omega\subset\mathbb{R}^n\) be a spatial domain with boundary \(\partial\Omega\), and let \(\mathcal{T}\) denote the time interval. A time-dependent partial differential equation can be written abstractly as
\begin{align}
    \mathcal{F}[u](\mathbf{x},t)&=0,
    &&(\mathbf{x},t)\in\Omega\times\mathcal{T}, \label{(1)}\\
    \mathcal{B}[u](\mathbf{x},t)&=0,
    &&(\mathbf{x},t)\in\partial\Omega\times\mathcal{T}, \label{(2)}\\
    \mathcal{I}[u](\mathbf{x},0)&=0,
    &&\mathbf{x}\in\Omega,
\end{align}
where $\mathcal{F}$, $\mathcal{B}$, and $\mathcal{I}$ denote the governing, boundary, and initial-condition operators, respectively. Nonhomogeneous data can be incorporated into these operators without changing the discussion below.

A PINN represents $u$ by a neural network $u_\theta$ with parameters $\theta\in\mathbb{R}^p$. Given the PDE residual, boundary, and initial-condition collocation sets $\Omega_F$, $\Omega_B$, and $\Omega_I$, respectively, the parameters are obtained by minimizing
\begin{align}
    f(\theta)
    &=\lambda_F\mathcal{L}_F(\theta)
      +\lambda_B\mathcal{L}_B(\theta)
      +\lambda_I\mathcal{L}_I(\theta), \label{(3)}\\
    \mathcal{L}_F(\theta)
    &=\frac{1}{N_f}\sum_{(\mathbf{x},t)\in\Omega_F}
      \bigl|\mathcal{F}[u_\theta](\mathbf{x},t)\bigr|^2, \label{(4)}\\
    \mathcal{L}_B(\theta)
    &=\frac{1}{N_b}\sum_{(\mathbf{x},t)\in\Omega_B}
      \bigl|\mathcal{B}[u_\theta](\mathbf{x},t)\bigr|^2, \label{(5)}\\
    \mathcal{L}_I(\theta)
    &=\frac{1}{N_0}\sum_{(\mathbf{x},0)\in\Omega_I}
      \bigl|\mathcal{I}[u_\theta](\mathbf{x},0)\bigr|^2. \label{(6)}
\end{align}
The weights $\lambda_F$, $\lambda_B$, and $\lambda_I$ balance loss components that may have substantially different scales. Such imbalance, together with stiffness and the derivative structure induced by the PDE operators, contributes to the ill-conditioning commonly observed in PINN training~\cite{wang2021understanding,krishnapriyan2021characterizing,rathore2024challenges}.

After absorbing the loss weights and normalization factors into the residual definitions, the objective can be written compactly as
\begin{equation}
    f(\theta)=\frac{1}{N_r}\sum_{i=1}^{N_r}R_i(\theta)^2,
    \label{eq:pinn-residual-form}
\end{equation}
where the $R_i$ collects the PDE, boundary, and initial-condition residuals. 

To describe the local curvature of this objective, suppose that each \(R_i:\mathbb{R}^p\to\mathbb{R}\) is twice continuously differentiable, and define
\[
    J_i(\theta):=\nabla R_i(\theta),
    \qquad
    Q_i(\theta):=\nabla^2R_i(\theta).
\]
For any direction \(h\in\mathbb{R}^p\),
\begin{equation}
    \nabla^2 f(\theta)[h,h]
    =
    \frac{2}{N_r}
    \sum_{i=1}^{N_r}
    \left[
        \bigl(J_i(\theta)^\top h\bigr)^2
        +
        R_i(\theta)h^\top Q_i(\theta)h
    \right].
    \label{eq:residual-hessian}
\end{equation}
The first term is nonnegative and corresponds to the Gauss–Newton curvature. The second term depends on the current residual and may have either sign. Consequently, although the PINN objective is a sum of squares, its full parameter-space Hessian need not be positive semidefinite. It can also be nearly singular or poorly conditioned, particularly when the two curvature components have substantially different scales.

This decomposition also determines the curvature information observed by a secant-based optimizer. The gradient displacement generated by the full PINN objective combines the nonnegative Gauss--Newton curvature with the residual-dependent curvature. Consequently, the raw secant pair may exhibit weak, sign-indefinite, or strongly anisotropic curvature even while the loss continues to decrease.

Self-scaling adapts the magnitude of the inverse approximation to the observed curvature scale. A positive curvature shift serves a complementary role by changing the local geometry represented by the secant displacement. This combination is especially relevant during high-accuracy refinement, where further progress depends on extracting stable curvature information from a small-residual but highly ill-conditioned objective. Section~\ref{subsec:weak-sc} develops the corresponding positive local-metric perspective from weak self-concordance, and Section~\ref{sec:motivation} realizes it within SSBroyden.

\subsection{Quasi-Newton Methods for PINN Training}
\label{sec:qn-prelim}
First-order methods such as Adam are commonly used during the early stage of PINN training, while curvature-based methods are often employed to obtain higher accuracy during refinement. Recent studies have considered BFGS and L-BFGS, self-scaled quasi-Newton methods, Gauss-Newton methods,  Levenberg-Marquardt methods, natural-gradient methods, and structured preconditioners~\cite{rathore2024challenges,urban2025unveiling,jnini2026curvature,kiyani2025optimizing,jnini2025gauss,muller2023achieving,jnini2025dual,levenberg1944method,dai2026tinns,alizadeh2025physics,shahab2026physics}. These methods represent curvature in different ways. Here we focus on secant-based quasi-Newton updates.

Let
\[
    g_k:=\nabla f(\theta_k),
    \qquad
    d_k:=-H_k g_k,
\]
where $H_k\succ0$ approximates an inverse local curvature metric. After choosing a step length $\alpha_k>0$ and setting $\theta_{k+1}=\theta_k+\alpha_k d_k$, the associated step and gradient displacement are
\[
    s_k:=\theta_{k+1}-\theta_k,
    \qquad
    y_k:=g_{k+1}-g_k.
\]
An inverse quasi-Newton update is constructed to satisfy the inverse secant equation
\begin{equation}
    H_{k+1}y_k=s_k.
    \label{eq:inverse-secant}
\end{equation}
Thus, the pair \((s_k,y_k)\) allows the method to learn local curvature from successive gradients without explicitly evaluating the Hessian.

Positive-definite quasi-Newton updates typically require the curvature condition $s_k^\top y_k>0$. A Wolfe line search helps enforce this condition, while damping or update skipping can be used when the observed secant pair is numerically unreliable~\cite{nocedal2006numerical,li2001modified}.

For example, the inverse BFGS update is
\begin{equation}
\begin{aligned}
    H_{k+1}^{\mathrm{BFGS}}
    =(I-\rho_ks_ky_k^\top)H_k(I-\rho_ky_ks_k^\top)
      +\rho_ks_ks_k^\top, \qquad   \rho_k&:=\frac{1}{s_k^\top y_k}.
\end{aligned}
\label{eq:inverse-bfgs}
\end{equation}
BFGS and L-BFGS are attractive for PINN training because they extract curvature information from gradient differences without explicitly forming the Hessian. L-BFGS furtherreduces memory requirements by retaining only a limited number of recent secant pairs.

Self-scaled Broyden methods additionally rescale the inverse approximation to account for changes in curvature magnitude. Define
\[
    r_k:=y_k^\top H_k y_k,
    \qquad
    v_k:=\frac{s_k}{s_k^\top y_k}
    -\frac{H_k y_k}{y_k^\top H_k y_k}.
\]
A representative inverse self-scaled Broyden update is
\begin{equation}
    H_{k+1}
    =\frac{1}{\tau_k}
    \left[
      H_k-
      \frac{H_k y_ky_k^\top H_k}{y_k^\top H_k y_k}
      +\phi_kr_kv_kv_k^\top
    \right]
    +\frac{s_ks_k^\top}{s_k^\top y_k},
    \label{eq:self-scaled-broyden}
\end{equation}
where $\tau_k>0$ is the self-scaling parameter and $\phi_k$ selects a member of the Broyden family. Setting $\tau_k=1$ recovers the standard inverse Broyden family, with particular choices of \(\phi_k\) yielding familiar updates such as BFGS.

Both BFGS and self-scaled Broyden methods therefore construct their curvature models from the observed secant pair \((s_k,y_k)\). For the PINN objective in~\eqref{eq:pinn-residual-form}, the gradient displacement \(y_k\) reflects both the nonnegative Gauss--Newton curvature and the residual-dependent curvature described in~\eqref{eq:residual-hessian}. Self-scaling adjusts the magnitude of the inverse approximation, but the secant target remains the raw displacement \(y_k\). Consequently, weak or poorly conditioned curvature in the PINN loss is inherited directly by the update.

Positive curvature shifts have previously been represented in quasi-Newton methods through modified secant displacements, including regularized L-BFGS and more general regularized limited-memory frameworks~\cite{tankaria2022regularized,kanzow2023regularization}. This shifted-secant machinery supplies a Hessian-free way to change the metric represented by the update. The next subsection develops the positive local metric and the associated step scale from weak self-concordance.

\subsection{Weak Self-Concordance and Regularized Newton}
\label{subsec:weak-sc}
Self-concordance controls how rapidly the local curvature of a function can change. 
For a \(C^3\) function \(f\), we use the directional notation
\[
    \nabla^2f(\theta)[h,h]
    :=
    \left.
    \frac{d^2}{dt^2}f(\theta+t h)
    \right|_{t=0},
    \qquad
    \nabla^3f(\theta)[h,h,h]
    :=
    \left.
    \frac{d^3}{dt^3}f(\theta+t h)
    \right|_{t=0}.
\]
A convex function \(f\) is \(\kappa\)-self-concordant if
\begin{equation}
    \left|
        \nabla^3 f(\theta)[h,h,h]
    \right|
    \le
    2\kappa
    \left(
        \nabla^2 f(\theta)[h,h]
    \right)^{3/2},
    \qquad
    \forall\theta,h.
    \label{eq:classical-sc}
\end{equation}
Thus, the third-order variation along \(h\) is controlled by the curvature measured along the same direction. 

For a nonconvex function, however, \(\nabla^2 f(\theta)[h,h]\) may be negative and therefore cannot directly serve as a positive local scale. Goldfarb et al.~\cite{goldfarb2025non} address this issue through weak self-concordance. For \(\ell>0\), define
\[
    f_\ell(\theta)
    :=
    f(\theta)+\frac{\ell}{2}\|\theta\|^2.
\]
The function \(f\) is called \((\kappa,\ell)\)-weakly self-concordant when \(f_\ell\) is \(\kappa\)-self-concordant and
\[
    \nabla^2 f_\ell(\theta)
    =
    \nabla^2 f(\theta)+\ell I
    \succ 0.
\]
Since the quadratic term has zero third derivative, the corresponding self-concordant condition is
\begin{equation}
\left|
\nabla^3 f(\theta)[h,h,h]
\right|
\le
2\kappa
\left(
\nabla^2 f(\theta)[h,h]
+
\ell\|h\|^2
\right)^{3/2},
\qquad
\forall\theta,h .
\end{equation} 
The shifted matrix \(M_\ell(\theta):=\nabla^2 f(\theta)+\ell I\) defines a positive local curvature metric. Goldfarb et al.~\cite{goldfarb2025non} use this metric in a regularized Newton method. At a point \(\theta\), the the direction is obtained by minimizing the regularized quadratic model
\begin{equation}
\begin{aligned}
    d_\ell(\theta)
    :=
    \arg\min_d\;
    \bigg\{
        \nabla f(\theta)^\top d
        +
        \frac{1}{2}d^\top\nabla^2 f(\theta)d
        +
        \frac{\ell}{2}\|d\|^2
    \bigg\}=-\bigl(\nabla^2 f(\theta)+\ell I\bigr)^{-1}\nabla f(\theta).
\end{aligned}
\label{eq:regularized-newton-model}
\end{equation}
The shift \(\ell I\) supplies a positive curvature offset to the local quadratic model. Notice that the direction still uses the gradient \(\nabla f(\theta)\) of the original objective. Thus, the shift regularizes the curvature model without changing the optimization target.

The shifted metric also defines the regularized Newton decrement
\begin{equation}
    \lambda_\ell(\theta)
    :=
    \sqrt{
        \nabla f(\theta)^\top
        (\nabla^2 f(\theta)+\ell I )^{-1}
        \nabla f(\theta)
    }.
    \label{eq:shifted-decrement}
\end{equation}
The corresponding self-concordant damped step size is
\begin{equation}
    \alpha_{\mathrm{sc}}
    =
    \frac{1}{1+\kappa\lambda_\ell(\theta)} .
    \label{eq:goldfarb-sc-step}
\end{equation}

Weak self-concordance therefore provides two linked geometric quantities: the positive shifted metric \(M_\ell(\theta)\) and the decrement \(\lambda_\ell(\theta)\) measured in that metric. The shift makes curvature-relative control well defined for a nonconvex objective, while the decrement calibrates the step using the same local geometry.

Applied directly, this construction requires the shifted Hessian and its inverse action. Quasi-Newton methods instead learn a local metric from successive steps and gradient displacements. Curvature-adaptive step rules based on self-concordant geometry have also been developed for quasi-Newton directions~\cite{gao2019quasi}. Section~\ref{sec:motivation} combines these ingredients by representing the positive shifted metric through the secant information used by SSBroyden and approximating the corresponding step scale with the learned inverse metric.

\section{Motivation and the \textsc{SCORE} Method}
\label{sec:motivation}

The preceding sections lead to a direct construction. The residual-dependent component of the PINN Hessian can make the raw secant curvature weak, indefinite, or poorly conditioned. Self-scaled Broyden adapts the scale of the inverse approximation while continuing to learn from this raw secant displacement. Weak self-concordance supplies the two additional ingredients needed here: a positive shifted local metric and a decrement-based step measured in that metric. We now represent the shifted metric through the SSBroyden secant pair and realize the corresponding decrement using the learned inverse approximation.

\subsection{From a Positive Local Metric to a Shifted Secant Pair}
\label{sec:shifted-secant-motivation}

Recall the accepted step and gradient displacement
\[
    s_k:=\theta_{k+1}-\theta_k,
    \qquad
    y_k:=\nabla f(\theta_{k+1})-\nabla f(\theta_k).
\]
By the fundamental theorem of calculus,
\begin{equation}
    y_k
    =
    \int_0^1
    \nabla^2 f(\theta_k+t s_k)s_k\,dt
    =
    \left(
        \int_0^1
        \nabla^2 f(\theta_k+t s_k)\,dt
    \right)s_k .
    \label{eq:secant-average-hessian}
\end{equation}
Thus, \(y_k\) is the action of the averaged Hessian along the accepted step.

At iteration \(k\), consider the positively shifted local metric
\[
    \nabla^2 f(\theta)+\mu_k I,
    \qquad
    \mu_k\geq0.
\]
Holding \(\mu_k\) fixed along the accepted segment, its secant action is
\begin{equation}
\begin{aligned}
    \int_0^1
    \left(
        \nabla^2 f(\theta_k+t s_k)+\mu_k I
    \right)s_k\,dt
    &=
    y_k+\mu_k s_k .
\end{aligned}
\label{eq:shifted-secant-integral}
\end{equation}
This gives the shifted gradient displacement
\begin{equation}
    \widetilde y_k
    :=
    y_k+\mu_k s_k .
    \label{eq:shifted-secant}
\end{equation}
The pair \((s_k,\widetilde y_k)\) represents the action of the averaged shifted metric along the accepted step, and the corresponding inverse secant equation is
\begin{equation}
    H_{k+1}\widetilde y_k=s_k .
    \label{eq:shifted-inverse-secant}
\end{equation}
Its directional curvature satisfies
\[
    s_k^\top\widetilde y_k
    =
    s_k^\top y_k+\mu_k\|s_k\|^2 .
\]
Hence, the shift preserves the observed gradient displacement and adds a positive curvature offset in the direction used by the secant update. Replacing \(y_k\) by \(\widetilde y_k\) in the self-scaled Broyden formula makes the inverse approximation learn this shifted local geometry without explicitly forming the Hessian.

The remaining quantities are the shift magnitude \(\mu_k\) and the step length used to generate \(s_k\). Both are determined from the current learned metric.

\subsection{Adaptive Quasi-Newton Realization}
\label{sec:algorithms}

Let \(H_k\succ0\) denote the inverse curvature approximation constructed from the previous shifted secant pairs, and define
\[
    g_k:=\nabla f(\theta_k),
    \qquad
    d_k:=-H_k g_k .
\]
The size of the full quasi-Newton step in the learned metric is measured by
\begin{equation}
    \lambda_k
    :=
    \sqrt{\max\{g_k^\top H_k g_k,0\}} .
    \label{eq:qn-decrement}
\end{equation}
When \(H_k\succ0\), the maximum is unnecessary in exact arithmetic and is included as a numerical safeguard. Since \(d_k=-H_k g_k\),
\[
    \lambda_k^2
    =
    -g_k^\top d_k
    =
    d_k^\top H_k^{-1}d_k .
\]
Thus, \(\lambda_k\) measures the proposed full step in the local metric currently represented by \(H_k\).

Motivated by the self-concordant damping rule in~\eqref{eq:goldfarb-sc-step}, we use the normalized candidate
\begin{equation}
    \alpha_k^{\mathrm{sc}}
    :=
    \frac{1}{1+\lambda_k}.
    \label{eq:sc-step}
\end{equation}
The candidate approaches a unit step when the learned-metric displacement is small and decreases as that displacement grows. It is accepted when it satisfies the strong Wolfe conditions; otherwise, a standard strong Wolfe line search is performed along \(d_k\).

The same decrement determines the positive shift used in the subsequent secant update:
\begin{equation}
    \mu_k
    :=
    \operatorname{clip}
    \bigl(
        \eta\lambda_k,[\mu_{\min},\mu_{\max}]
    \bigr)
    :=
    \min\{
        \max\{\eta\lambda_k,\mu_{\min}\},
        \mu_{\max}
    \},
    \label{eq:adaptive-shift}
\end{equation}
where \(\eta>0\) and \(0<\mu_{\min}\leq\mu_{\max}\). A large decrement produces a smaller candidate step and a stronger curvature shift. As the decrement decreases, the candidate approaches a full step and the shift approaches \(\mu_{\min}\).

Equations~\eqref{eq:sc-step} and~\eqref{eq:adaptive-shift} form the central coupling of \textsc{SCORE}: the metric learned at iteration \(k\) controls both the displacement proposed with the current metric and the stabilization used to construct the next metric. The iteration therefore follows
\[
    H_k
    \;\longrightarrow\;
    (d_k,\lambda_k)
    \;\longrightarrow\;
    (\alpha_k^{\mathrm{sc}},\mu_k)
    \;\longrightarrow\;
    (s_k,\widetilde y_k)
    \;\longrightarrow\;
    H_{k+1}.
\]
Algorithm~\ref{alg:score} summarizes the resulting procedure.

\begin{algorithm}[!ht]
\caption{\textsc{SCORE}: Self-Concordance-Inspired Quasi-Newton Method with Shifted Secant Geometry}
\label{alg:score}
\KwIn{
    \(\theta_0\),
    \(H_0\succ0\),
    number of iterations \(T\),
    \(\eta>0\),
    \(0<\mu_{\min}\le\mu_{\max}\),
    Wolfe parameters \(0<c_1<c_2<1\)
}
\KwOut{\(\theta_T\)}

\(f_0\gets f(\theta_0)\)\;
\(g_0\gets\nabla f(\theta_0)\)\;

\For{\(k=0,1,\ldots,T-1\)}{
    \(d_k\gets-H_k g_k\)
    \tcp*[r]{Quasi-Newton direction}

    \(\lambda_k
    \gets
    \sqrt{\max\{g_k^\top H_k g_k,0\}}\)
    \tcp*[r]{Quasi-Newton decrement}

    \(\alpha_k^{\mathrm{sc}}
    \gets
    (1+\lambda_k)^{-1}\)
    \tcp*[r]{Candidate step length}

    \(\widehat\theta_k
    \gets
    \theta_k+\alpha_k^{\mathrm{sc}}d_k\)\;

    \(\widehat f_k\gets f(\widehat\theta_k)\)\;

    \(\widehat g_k\gets\nabla f(\widehat\theta_k)\)\;

    \eIf{
        \(\widehat f_k
        \le
        f_k+c_1\alpha_k^{\mathrm{sc}}g_k^\top d_k\)
        \textnormal{ and }
        \(
        |\widehat g_k^\top d_k|
        \le
        c_2|g_k^\top d_k|
        \)
    }{
        \(\alpha_k\gets\alpha_k^{\mathrm{sc}}\)\;

        \(\theta_{k+1}\gets\widehat\theta_k\)\;

        \(f_{k+1}\gets\widehat f_k\)\;

        \(g_{k+1}\gets\widehat g_k\)\;
    }{
        \((\alpha_k,f_{k+1},g_{k+1})
        \gets
        \operatorname{WolfeLineSearch}
        (f,\theta_k,d_k,g_k,f_k,c_1,c_2)\)
        \tcp*[r]{Wolfe line search}

        \(\theta_{k+1}
        \gets
        \theta_k+\alpha_kd_k\)\;
    }
    \(\mu_k
    \gets
    \operatorname{clip}
    (\eta\lambda_k,[\mu_{\min},\mu_{\max}])\)
    \tcp*[r]{Adaptive curvature shift}

    \(s_k\gets\theta_{k+1}-\theta_k\)\;

    \(y_k\gets g_{k+1}-g_k\)\;

    \(\widetilde y_k\gets y_k+\mu_k s_k\)
    \tcp*[r]{Shifted secant displacement}

    \(H_{k+1}
    \gets
    \operatorname{SSBroydenUpdate}
    (H_k,s_k,\widetilde y_k)\)
    \tcp*[r]{Safeguarded inverse update}
}

\Return{\(\theta_T\)}\;
\end{algorithm}

The loss value, gradient, and Wolfe conditions in Algorithm~\ref{alg:score} are all evaluated using the original PINN objective \(f\). The positive shift is introduced only in the secant pair used to update the local inverse curvature approximation. Thus, \textsc{SCORE} regularizes the learned curvature model without changing the underlying training objective.

\section{Experiments}
\label{sec:experiments}

We evaluate SCORE on four benchmark PDEs: the viscous Burgers, Kuramoto–Sivashinsky, Korteweg–de Vries, and complex Ginzburg–Landau equations. We compare SCORE with BFGS and self-scaled Broyden (SSBroyden), both implemented with the same Wolfe line search. Unless stated otherwise, all methods use the same network architecture, parameter initialization, Adam warm start, collocation-sampling schedule, outer–inner refinement budget, Wolfe parameters, and evaluation set. Thus, the method-specific differences are limited to the quasi-Newton update and, for SCORE, the shifted secant displacement and self-concordant candidate-step test.

For each problem, we generate a fixed test set of 90,000 randomly sampled points and use it to evaluate all three optimizers. These test points are sampled independently of the training points and are not included in the training batches. We use $N_f$, $N_b$, and $N_0$ to denote the numbers of points associated with the PDE residual, boundary condition, and initial condition, respectively, whenever these terms are present.

During quasi-Newton refinement, each outer block begins with a newly sampled training batch. The optimizer then performs a fixed maximum number of inner iterations on that batch. The inverse-Hessian approximation obtained at the end of one block is carried over to the next block. SCORE uses the same regularization strategy and hyperparameters across all four benchmarks, without problem-specific tuning. Unless stated otherwise, the networks are initialized using Glorot-normal initialization, and all computations are performed in double precision.

\begin{figure}[!htb]
    \centering
    % Row 1
    \begin{subfigure}[t]{0.49\textwidth}
        \captionsetup{skip=-1.2\baselineskip, margin=0pt}
        \label{fig:panelA}
        \includegraphics[width=\linewidth]{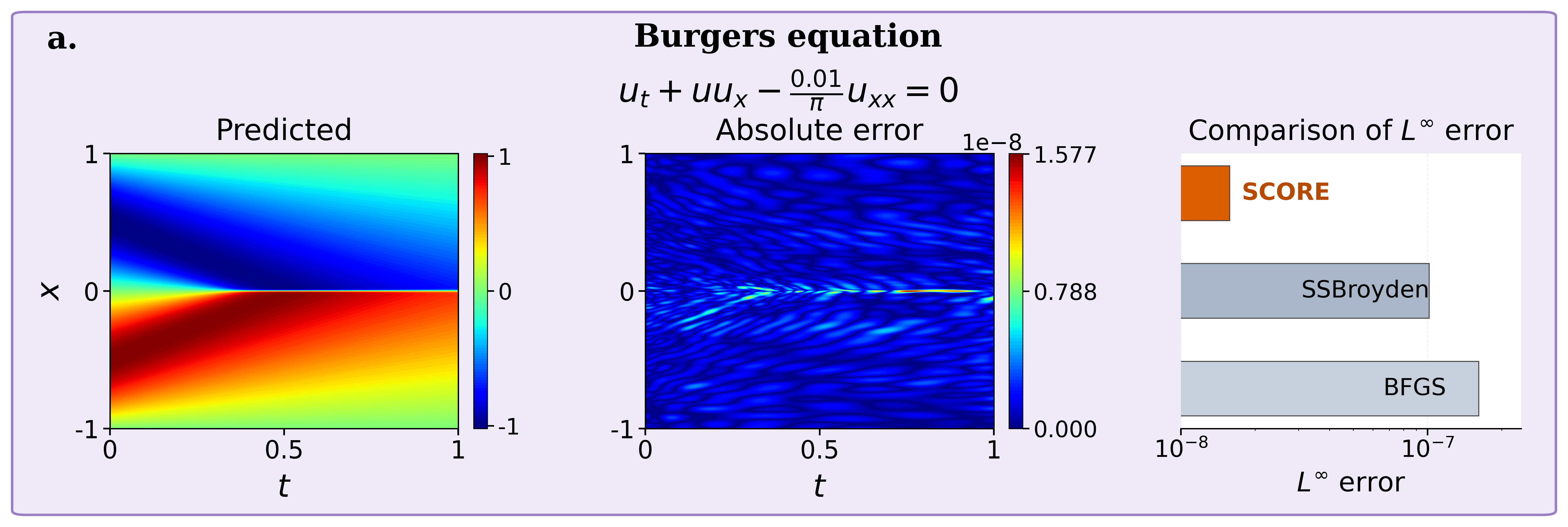}
    \end{subfigure}
    \hfill
    \begin{subfigure}[t]{0.49\textwidth}
        \captionsetup{skip=-1.2\baselineskip, margin=0pt}
        \label{fig:panelB}
        \includegraphics[width=\linewidth]{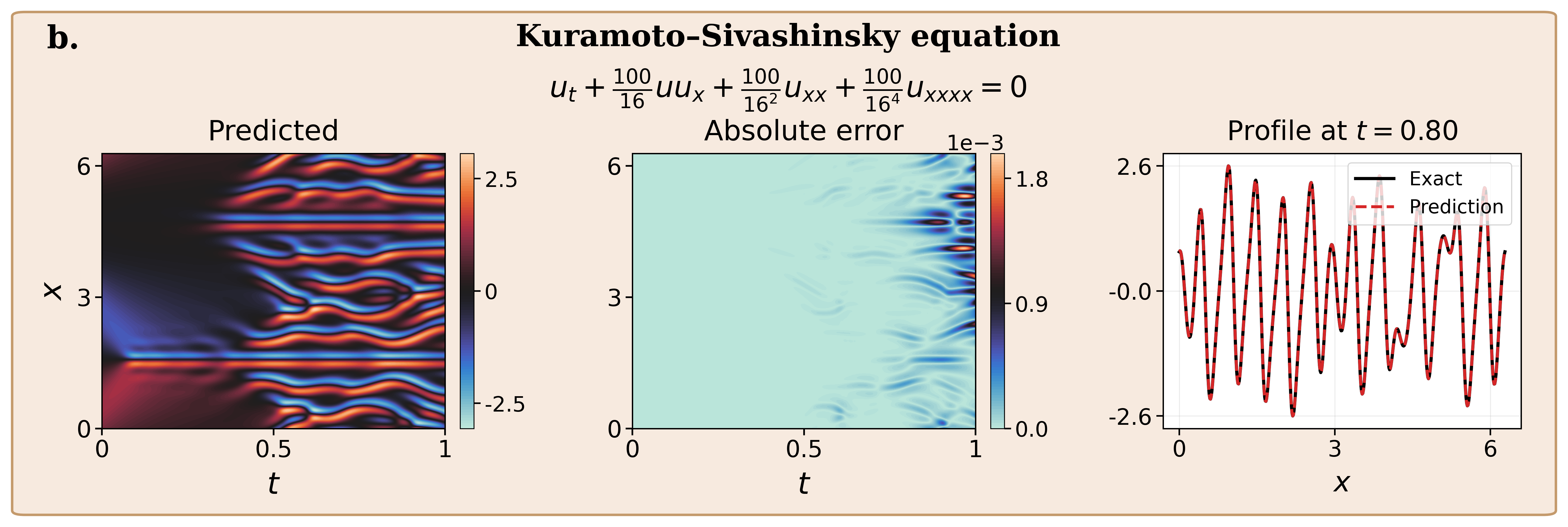}
    \end{subfigure}

    \vspace{0.8em}

    % Row 2
    \begin{subfigure}[t]{0.49\textwidth}
        \captionsetup{skip=-1.2\baselineskip, margin=0pt}
        \label{fig:panelC}
        \includegraphics[width=\linewidth]{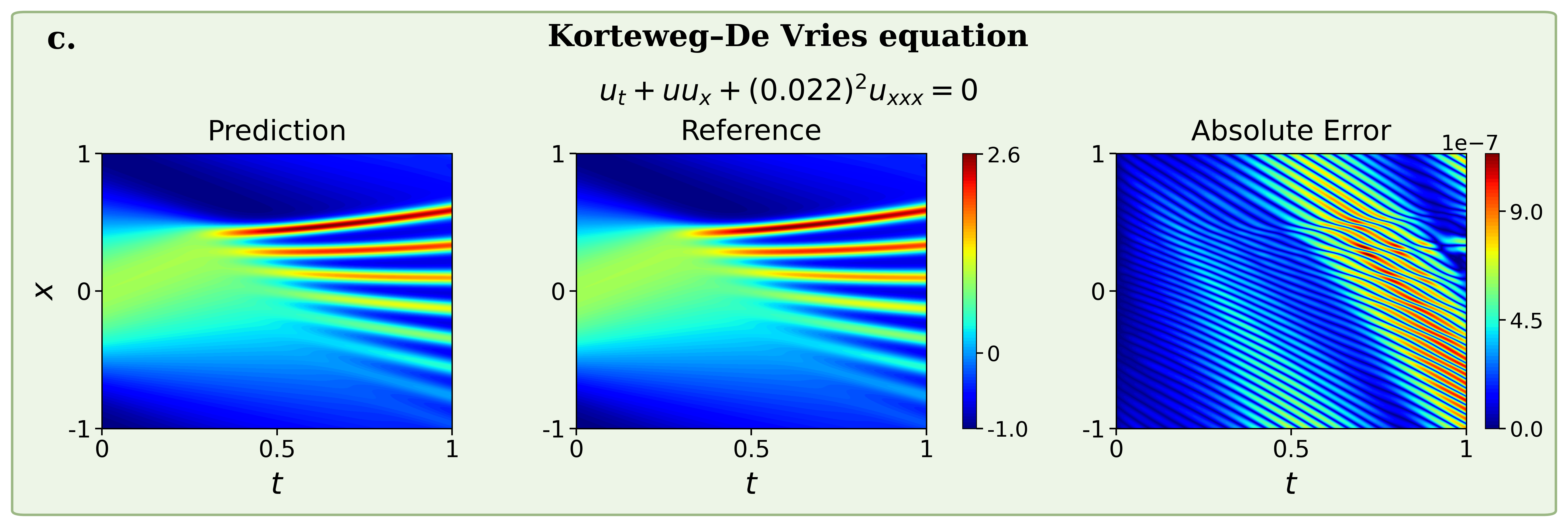}
    \end{subfigure}
    \hfill
    \begin{subfigure}[t]{0.49\textwidth}
        \captionsetup{skip=-1.2\baselineskip, margin=0pt}
        \label{fig:panelD}
        \includegraphics[width=\linewidth]{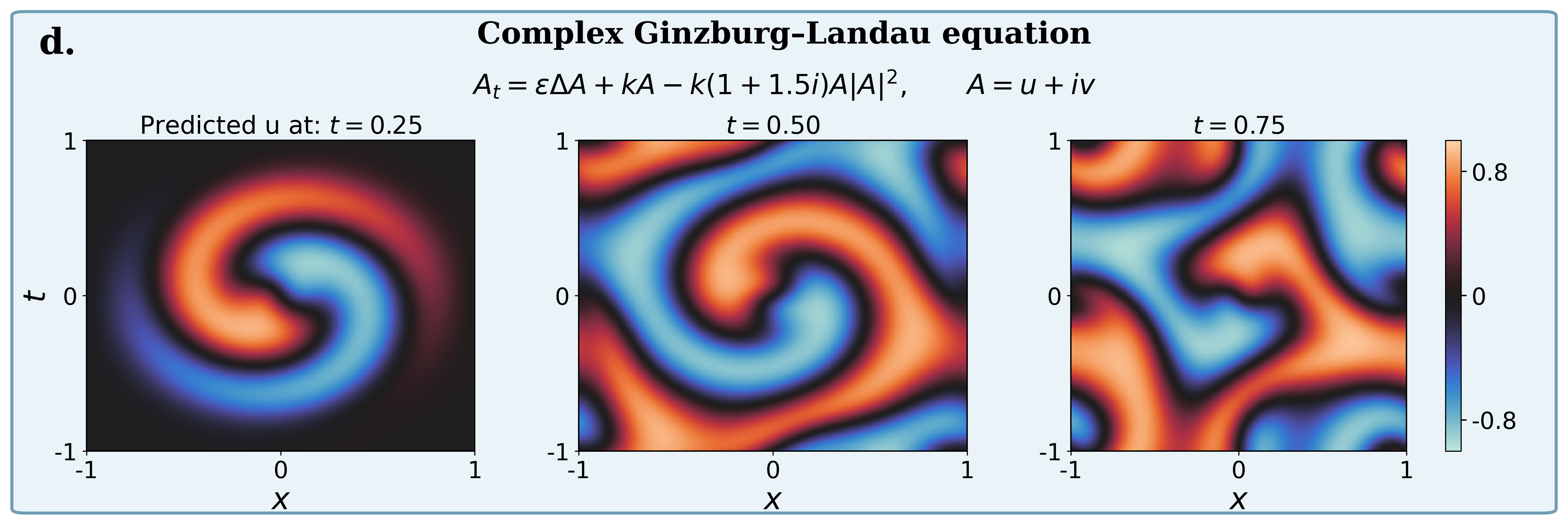}
    \end{subfigure}

    \vspace{0.8em}

    % Row 3
    \begin{subfigure}[t]{0.7\textwidth}
        \captionsetup{skip=-1.2\baselineskip, margin=0pt}
        \label{fig:panelE}
        \includegraphics[width=\linewidth]{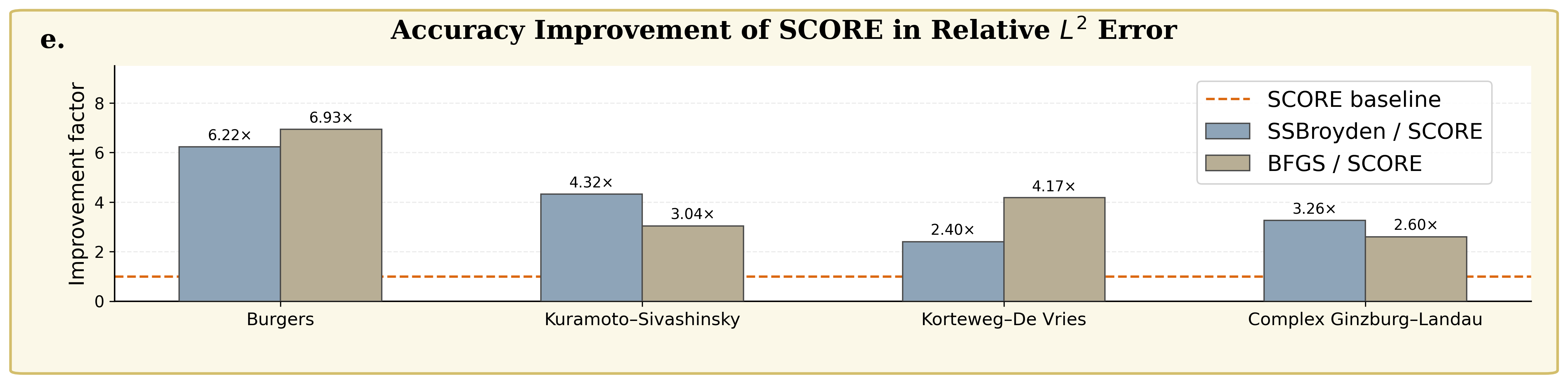}
    \end{subfigure}

    \caption{Overview of predictive performance on the four PDE benchmarks. Panels \textbf{A}--\textbf{D} show representative results for the Burgers, Kuramoto--Sivashinsky, Korteweg--de Vries, and complex Ginzburg--Landau equations, respectively, including prediction, pointwise absolute-error, and selected profile or snapshot comparisons where appropriate. 
    Panel~\textbf{E} reports the baseline-to-SCORE relative $L^2$-error ratios, $E_{\mbox{baseline}}/E_{\mbox{SCORE}}$. The dashed line at one denotes parity, and values greater than one favor SCORE.}\label{fig:overall}
\end{figure}

For a real-valued scalar solution, we report the relative $L^2$ and $L^\infty$ errors defined as 
\begin{align*}
     \text{Relative } L^2 \text{ error}&=\frac{\sqrt{\sum_{k=1}^{N}\left|\hat{u}(\mathbf{x}_k,t_k)-u(\mathbf{x}_k,t_k)\right|^2}}{\sqrt{\sum_{k=1}^{N}\left|u(\mathbf{x}_k,t_k)\right|^2}},\\
     L^{\infty} \text{ error}&=\max_{1 \leq k \leq N}\left|\hat{u}(\mathbf{x}_k,t_k)-u(\mathbf{x}_k,t_k)\right|.
\end{align*}
Here, $u$ denotes the reference solution, $\hat u$ denotes the PINN prediction, and $N$ is the number of test points. For the complex Ginzburg–Landau equation, a combined complex-field error is defined separately in \S\ref{sec:cgl}. Fig.~\ref{fig:overall} provides an overview of the four benchmark problems and summarizes the relative $L^2$ improvements obtained by SCORE.

For the Burgers benchmark, we also report the mean wall-clock time per quasi-Newton refinement block. Each block includes the sampling of a new training batch and up to 200 inner quasi-Newton iterations. To remove one-time just-in-time compilation and warm-up effects, we discard the first 10 blocks and compute the mean and standard deviation over blocks 11–100. The reported time excludes the Adam warm start, test-set evaluation, plotting, and post-block diagnostics.

\subsection{Viscous Burgers Equation}
\label{Sec Burgers}
We first consider the one-dimensional viscous Burgers equation, a standard benchmark for nonlinear convection–diffusion problems:
\begin{subequations}
\begin{align}
&u_t + u u_x = \frac{0.01}{\pi} u_{xx}, \quad x \in \Omega=[-1,1],\; t \in T=[0,1],\\
&u(-1,t) = u(1,t), \quad t \in T,\\
&u(x,0) = -\sin(\pi x), \quad x \in \Omega.
\end{align}
\end{subequations}

All three optimizers use the same fully connected neural network with three hidden layers, 30 neurons per layer, hyperbolic-tangent activations, and a scalar output. To impose spatial periodicity as a hard constraint, we transform the spatial coordinate using the periodic Fourier embedding
\[
\left[
\cos\left(\frac{2\pi x}{L_x}\right),\,
\sin\left(\frac{2\pi x}{L_x}\right),\,
\dots,\,
\cos\left(\frac{2\pi m x}{L_x}\right),\,
\sin\left(\frac{2\pi m x}{L_x}\right)
\right],
\]
with $L_x=2$ and $m=5$. The embedded spatial features are concatenated with the temporal coordinate before being passed to the network. Because the network depends on $x$ only through periodic features, its output is periodic in space by construction.

All methods are first trained with Adam for 5,000 epochs, using an initial learning rate of $10^{-3}$ and an exponential decay schedule. The quasi-Newton phase then consists of 750 refinement blocks, each allowing up to 200 inner iterations, for a nominal budget of 150,000 quasi-Newton iterations. At the beginning of every block, 30,000 collocation points are sampled uniformly from the space–time domain.

\begin{figure}[!htb]
\centering
   \begin{minipage}{0.45\textwidth}
     \centering
     \includegraphics[width=\linewidth]{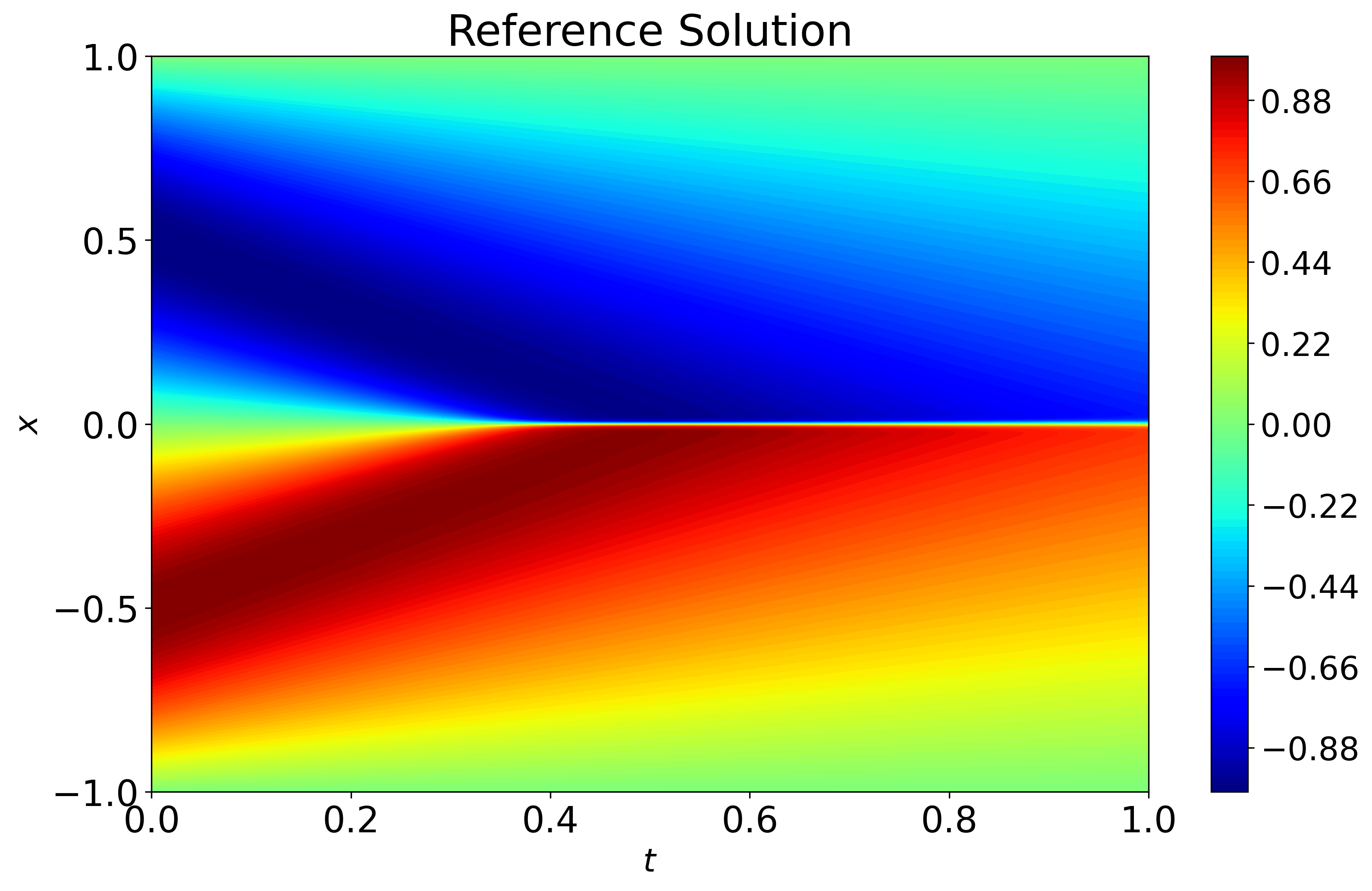} 
   \end{minipage}  
   \begin{minipage}{0.45\textwidth}
     \centering
     \includegraphics[width=\linewidth]{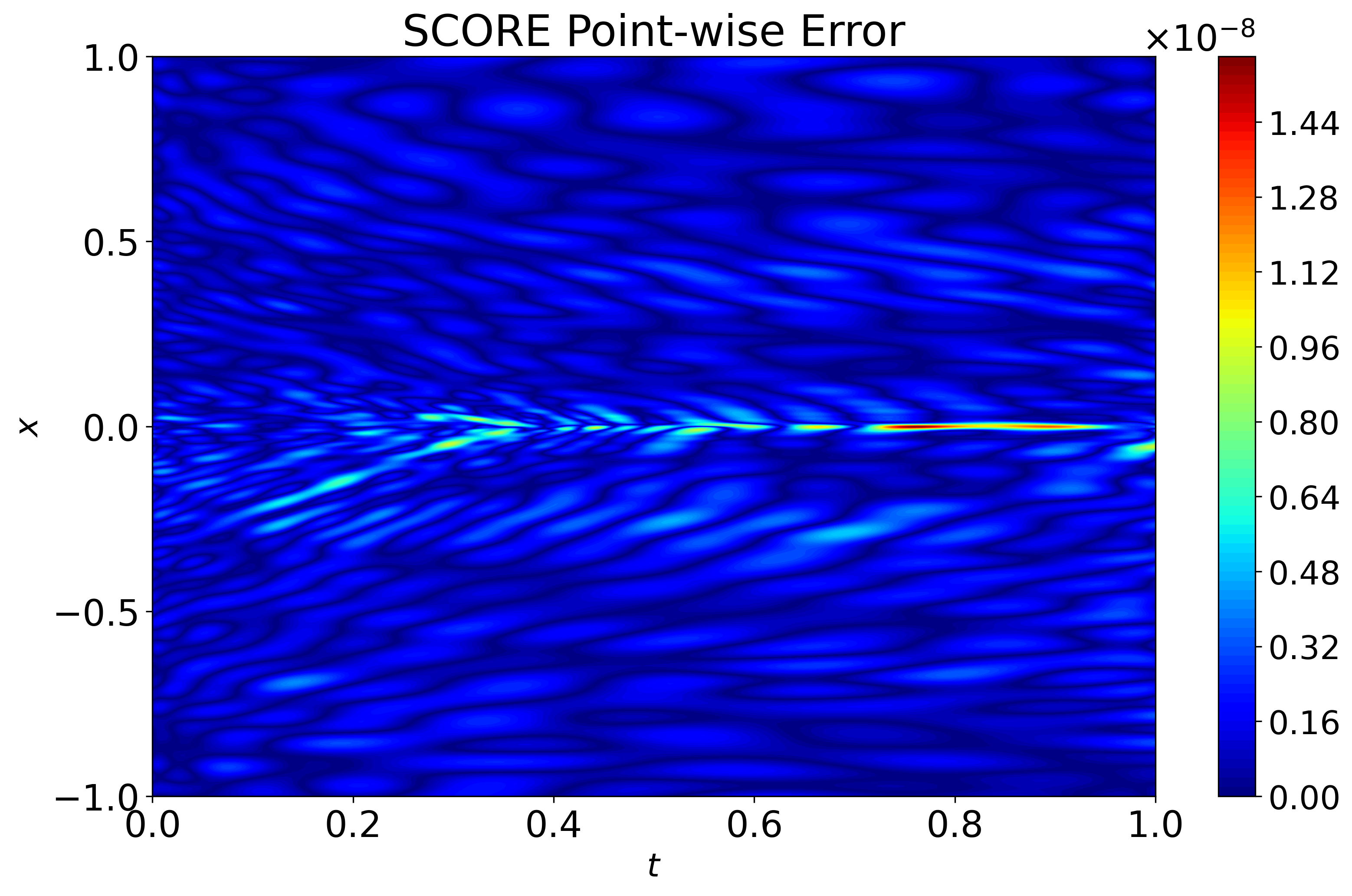} 
   \end{minipage} 

   \begin{minipage}{0.45\textwidth}
     \centering
     \includegraphics[width=\linewidth]{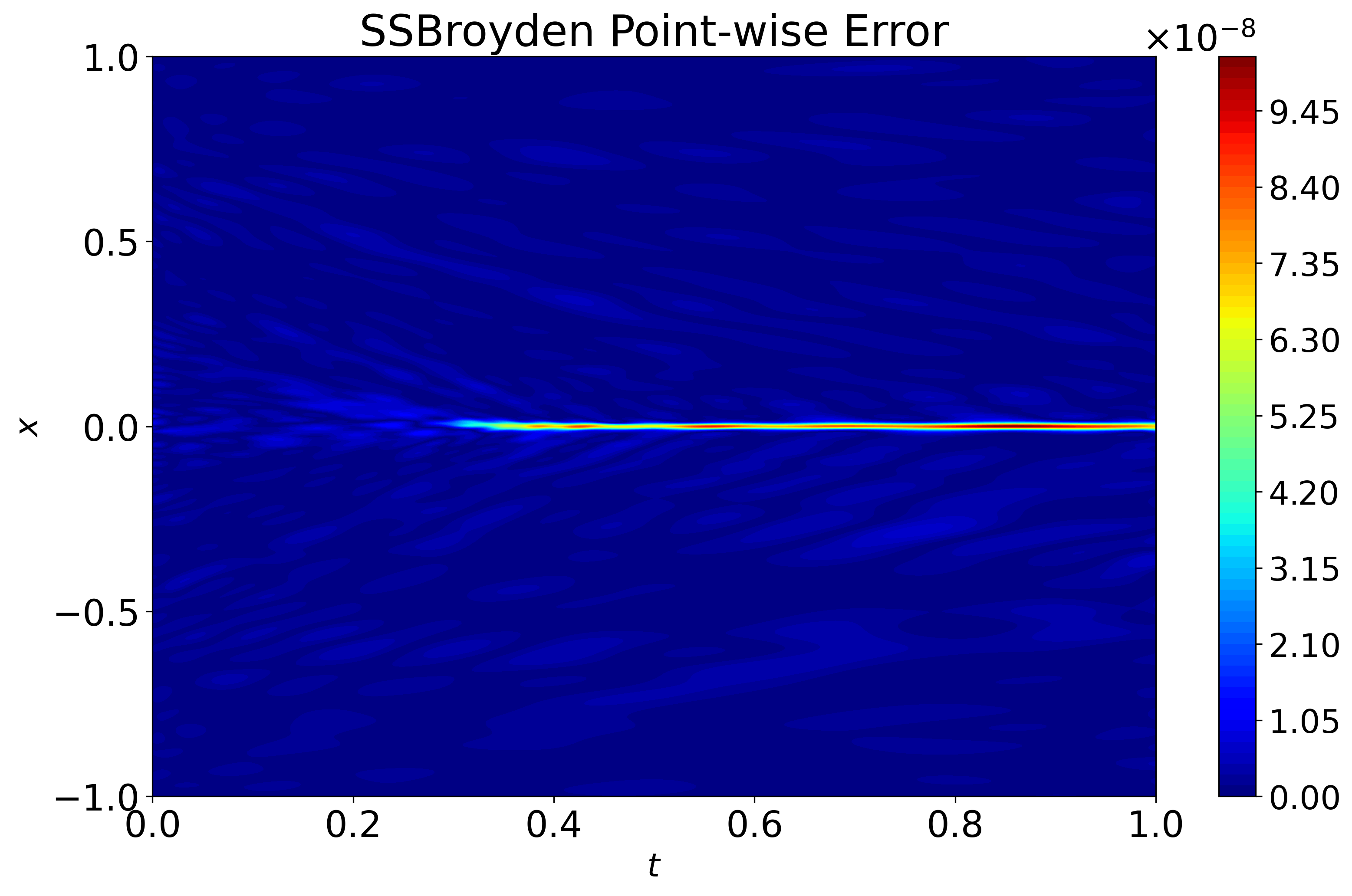} 
   \end{minipage}  
   \begin{minipage}{0.45\textwidth}
     \centering
     \includegraphics[width=\linewidth]{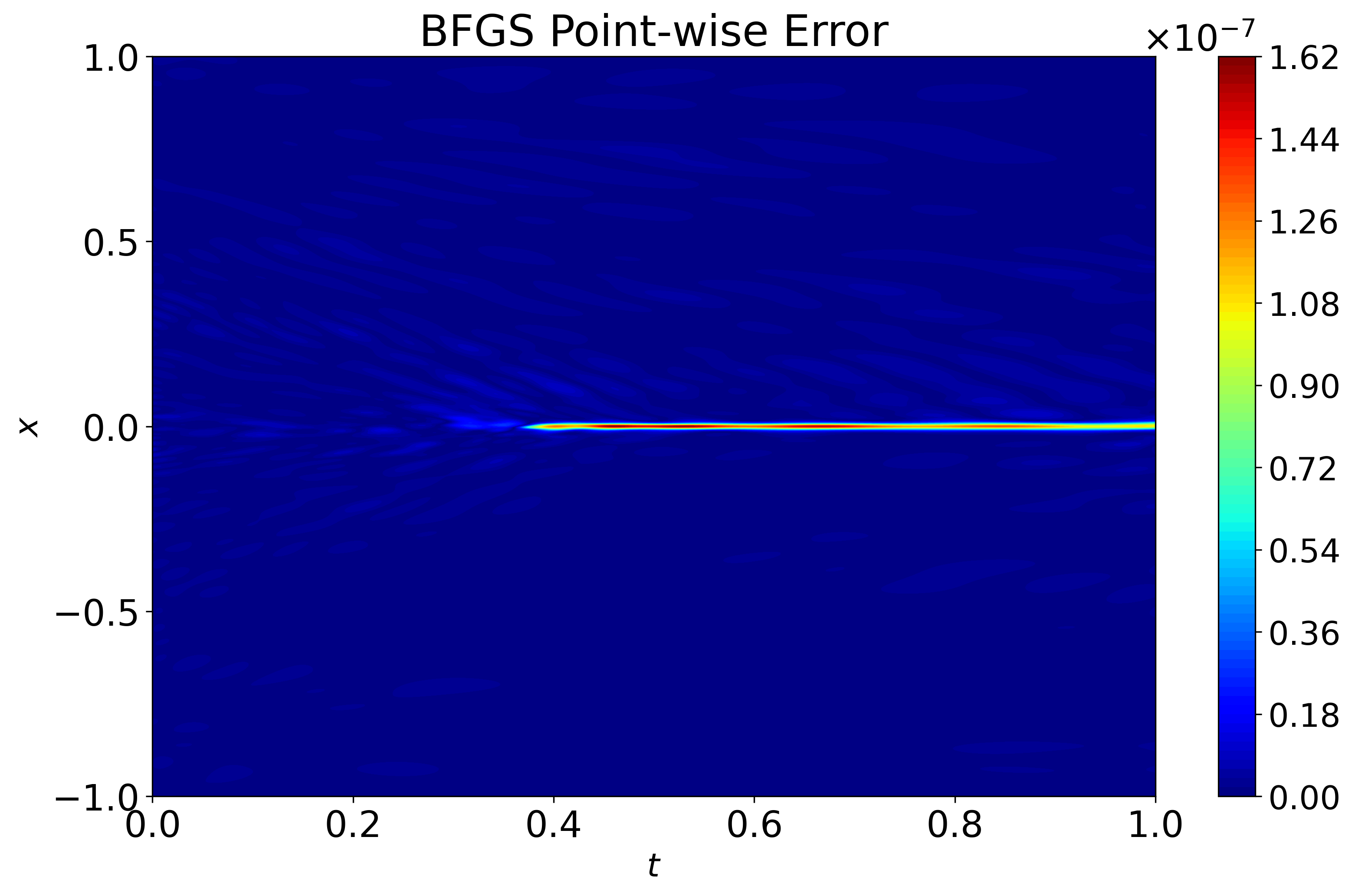} 
   \end{minipage} 
   
    \caption{{\bf Reference solution and pointwise absolute errors for the viscous Burgers equation.} The upper-left panel shows the reference solution, and the remaining panels show the absolute errors produced by SCORE, SSBroyden, and BFGS over the full space–time domain. Predicted fields are omitted because they are visually indistinguishable from the reference solution at the displayed resolution. Each error panel uses its own color scale, as indicated by the corresponding color bar; quantitative comparisons should therefore be based on Table 1. }\label{fig:burgers_vis}
\end{figure}

Fig.~~\ref{fig:burgers_vis} presents the reference solution and the pointwise absolute-error fields obtained with SCORE, SSBroyden, and BFGS. Because the predicted fields are visually indistinguishable from the reference solution at the displayed resolution, only the error fields are shown. With a common color normalization, the SCORE error is visibly smaller and more localized near the sharp transition region.

The numerical results in Table~\ref{tab:burgers_error} confirm this observation. SCORE achieves a relative $L^2$ error of $2.25\times 10^{-9}$, compared with $1.40\times 10^{-8}$ for SSBroyden and $1.56\times 10^{-8}$ for BFGS. These values correspond to reductions of approximately $6.2\times$ and $6.9\times$, respectively. SCORE also achieves the smallest $L^\infty$ error, indicating improved accuracy in the region of maximum pointwise discrepancy.

Meanwhile, the mean wall-clock times per refinement block are comparable across the three methods, indicating that the shifted secant construction and candidate-step test introduce little additional computational overhead in this implementation.

\begin{figure}[!htb]
\centering
   \begin{minipage}{0.6\textwidth}
     \centering
     \includegraphics[width=\linewidth]{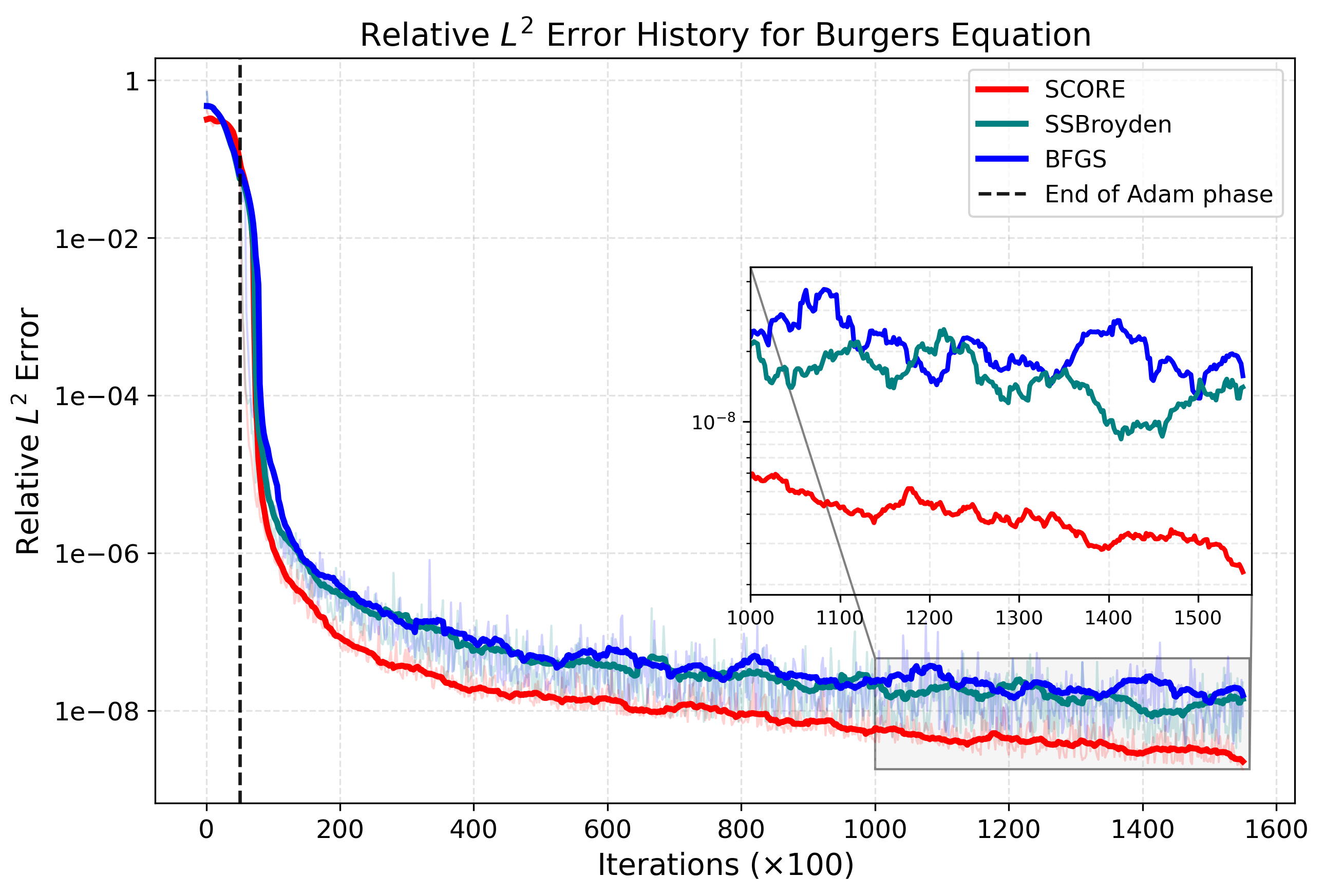} 
   \end{minipage}  
    \caption{{\bf Relative $L^2$ error history for the Burgers equation.} The dashed vertical line marks the end of the Adam warm-start phase and the beginning of quasi-Newton refinement. While all three methods reduce the error rapidly after the switch, \textsc{SCORE} continues to achieve more effective late-stage refinement. The inset zooms into the high-accuracy regime, where \textsc{SCORE} consistently attains lower error than BFGS and SSBroyden.}
\label{fig:burgers_hist}
\end{figure}

Fig.~\ref{fig:burgers_hist} shows the relative $L^2$ error throughout the Adam and quasi-Newton phases. All three methods rapidly reduce error after the transition from Adam. In the later stages, however, BFGS and SSBroyden approach a plateau, whereas SCORE continues to reduce the error and reaches a lower final level. The inset emphasizes this separation in the high-accuracy regime.

\begin{table}[t]
\centering
\caption{Quantitative comparison for the viscous Burgers equation. Lower error values indicate better accuracy. Time per block reports the mean wall-clock time per second-order optimizer block, with the standard deviation computed over blocks 11--100 after excluding the first 10 warm-up blocks. Each block includes batch resampling and up to $200$ inner quasi-Newton iterations, but excludes Adam warm-start, test evaluation, plotting, and post-block diagnostics.}
\label{tab:burgers_error}
\begin{tabular}{lccc}
\toprule
Optimizer & Relative $L^2$ Error & $L^\infty$ Error & Time / block (s)\\
\midrule
\textsc{SCORE}   & $\mathbf{2.25\times 10^{-9}}$ & $\mathbf{1.58\times 10^{-8}}$ & $1.25 \pm 1.11$\\
BFGS             & $1.56\times 10^{-8}$ & $1.62\times 10^{-7}$ & $1.43 \pm 0.93$\\
SSBroyden        & $1.40\times 10^{-8}$ & $1.02\times 10^{-7}$ & $1.34 \pm 1.02$\\
\bottomrule
\end{tabular}
\end{table}

These results are consistent with the motivation developed in \S~\ref{sec:shifted-secant-motivation}: the benefit of SCORE becomes most apparent when further progress depends on extracting reliable local curvature information rather than on obtaining coarse descent. The experiment shows that SCORE provides more effective late-stage refinement with the same training budget.

\subsubsection{Ablation and Mechanistic Interpretation}
\label{sec:burgers_ablation}

We further use the viscous Burgers equation to diagnose the two main components of \textsc{SCORE}: the shifted secant displacement \eqref{eq:shifted-secant} and the self-concordant candidate-step test \eqref{eq:sc-step}. This benchmark is a suitable diagnostic case because, after the Adam warm-start, the optimizer enters a high-accuracy refinement regime where further progress depends strongly on reliable curvature modeling and step-size selection.

Table~\ref{tab:burgers_ablation} isolates the contributions of the shifted secant displacement and the self-concordant candidate-step test. Removing either component increases the final relative $L^2$ error. Nevertheless, both the shift-only and candidate-step-only variants outperform the unmodified SSBroyden baseline, indicating that each component contributes independently to the improvement. Their combination produces the lowest error.

\begin{table}[t]
\centering
\caption{Ablation study on the viscous Burgers equation. The table isolates the two proposed components of \textsc{SCORE}: the shifted secant displacement and the self-concordant candidate-step test. Lower relative \(L^2\) error indicates better final accuracy.}
\label{tab:burgers_ablation}
\begin{tabular}{lccc}
\toprule
Method & Shifted secant \eqref{eq:shifted-secant} & SC test \eqref{eq:sc-step} & Relative \(L^2\) Error \(\downarrow\)\\
\midrule
Full \textsc{SCORE} & Yes & Yes & \(\mathbf{2.25\times 10^{-9}}\)\\
Shift only & Yes & No & \(5.32\times 10^{-9}\)\\
SC step only & No & Yes & \(9.99\times 10^{-9}\)\\
SSBroyden & No & No & \(1.40\times 10^{-8}\)\\
\bottomrule
\end{tabular}
\end{table}

The two components affect different stages of the quasi-Newton iteration. The shifted secant displacement replaces the raw gradient displacement \(y_k\) with
\[
    \widetilde y_k = y_k+\mu_k s_k,
\]
Consequently
\[
    s_k^\top \widetilde y_k
    =
    s_k^\top y_k+\mu_k\|s_k\|^2 .
\]
For \(\mu_k>0\) and \(s_k\neq 0\), the shift increases the curvature represented along the accepted step while retaining the directional information contained in $y_k$. The inverse quasi-Newton approximation is therefore updated toward the shifted local metric rather than toward the raw Hessian geometry alone.

The self-concordant candidate-step test acts earlier in the iteration. It proposes \(\alpha_k^{\rm sc}=(1+\lambda_k)^{-1}\) using the current quasi-Newton decrement and accepts the candidate only when the strong Wolfe conditions are satisfied. Otherwise, the method uses the same Wolfe line-search fallback as the baselines. Thus, the shifted displacement modifies the curvature information used in the inverse update, whereas the candidate-step test supplies an additional curvature-dependent proposal for the step length. The ablation results are consistent with these two mechanisms playing complementary roles.

\subsection{Kuramoto-Sivashinsky equation}
\label{sec:ks}
We next evaluate \textsc{SCORE} on the one-dimensional Kuramoto--Sivashinsky equation, a canonical nonlinear PDE for modeling spatio-temporal chaos and unstable dissipative dynamics. Owing to its nonlinear convective term together with competing destabilizing and stabilizing higher-order effects, this problem provides a challenging benchmark for PINNs on chaotic time-dependent systems~\cite{wang2024respecting}. The governing equation is given by
$$
\frac{\partial u}{\partial t}
+
\alpha u \frac{\partial u}{\partial x}
+
\beta \frac{\partial^2 u}{\partial x^2}
+
\gamma \frac{\partial^4 u}{\partial x^4}
= 0,
\qquad
t \in (0,1), \quad x \in (0,2\pi),
$$
subject to periodic boundary conditions, where $u=u(x,t)$ denotes the scalar solution field. In our experiments, we set
$\alpha=\frac{100}{16}, \beta=\frac{100}{16^2}, \gamma=\frac{100}{16^4}.$ The initial condition is chosen as
$$
u(x,0)=\cos(x)\bigl(1+\sin(x)\bigr).
$$
For the Kuramoto--Sivashinsky equation, we train the PINN to predict the scalar field $u(x,t)$ directly. The network is a fully connected MLP with three hidden layers of width 80 and a one-dimensional output. We normalize the temporal input within each training window and use a periodic Fourier feature embedding for the spatial coordinate $x$ to encode the periodic structure of the solution. In particular, for the spatial coordinate we use a Fourier feature embedding of degree $m=2$ and $L_x=2\pi$. These periodic spatial features are concatenated with the normalized time variable before being fed into the MLP.

To improve optimization over the long time horizon, we adopt a time-marching strategy based on temporal domain decomposition \cite{wang2024respecting,wang2025gradient,kiyani2025optimizing,krishnapriyan2021characterizing}. Specifically, the full time interval $[0,1]$ is divided into $20$ consecutive windows with time step $\Delta t = 0.05$, and a separate PINN is trained on each window. For the first window, the initial condition is taken from the reference solution data. For every subsequent window, the initial condition is set to the prediction of the trained model at the final time of the previous window, thereby propagating the solution forward in time. Within each window, we sample $30{,}000$ collocation points in the interior of the space-time subdomain and impose the initial condition on the full spatial grid through a soft penalty term with weight $\lambda_{\mathrm{IC}} = 500$.

Training is performed in two stages within each window. We warm up the optimization with Adam only in the first time window, using $5000$ epochs and a learning rate of $10^{-4}$ with exponential decay. After this warm-start phase, we switch to the second-order optimizers. For every window, including the first one, the second-order optimizers are run for $30{,}000$ inner iterations, with the stochastic batch refreshed every $200$ inner steps. To stabilize optimization, each new time window is initialized from the best parameters obtained in the previous window. This combination of temporal domain decomposition, warm-start transfer across windows, and second-order refinement yields a stable training pipeline for this chaotic time-dependent problem.

\begin{figure}[!htb]
\centering
   \begin{minipage}{0.45\textwidth}
     \centering
     \includegraphics[width=\linewidth]{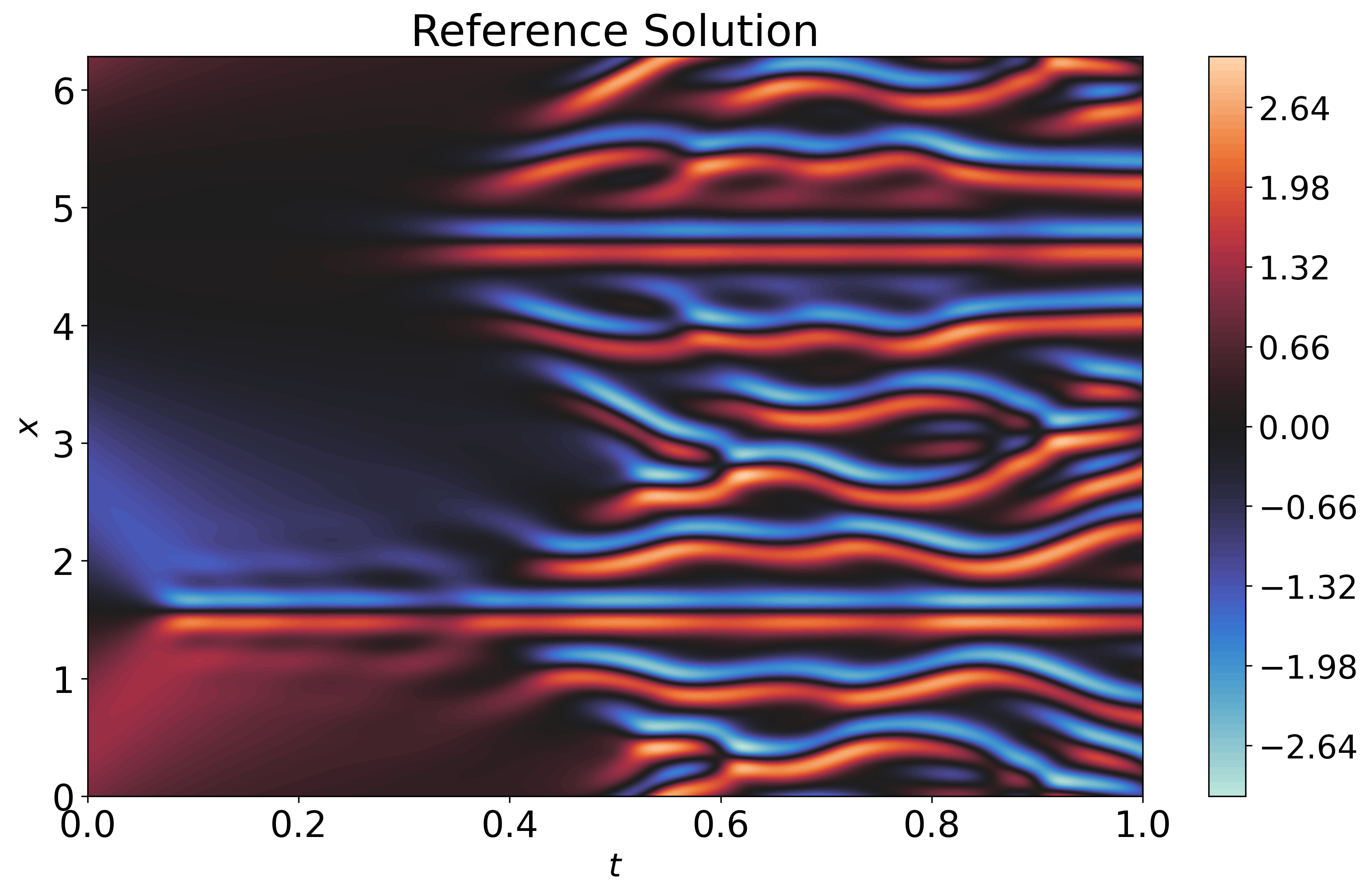} 
   \end{minipage}  
   \begin{minipage}{0.45\textwidth}
     \centering
     \includegraphics[width=\linewidth]{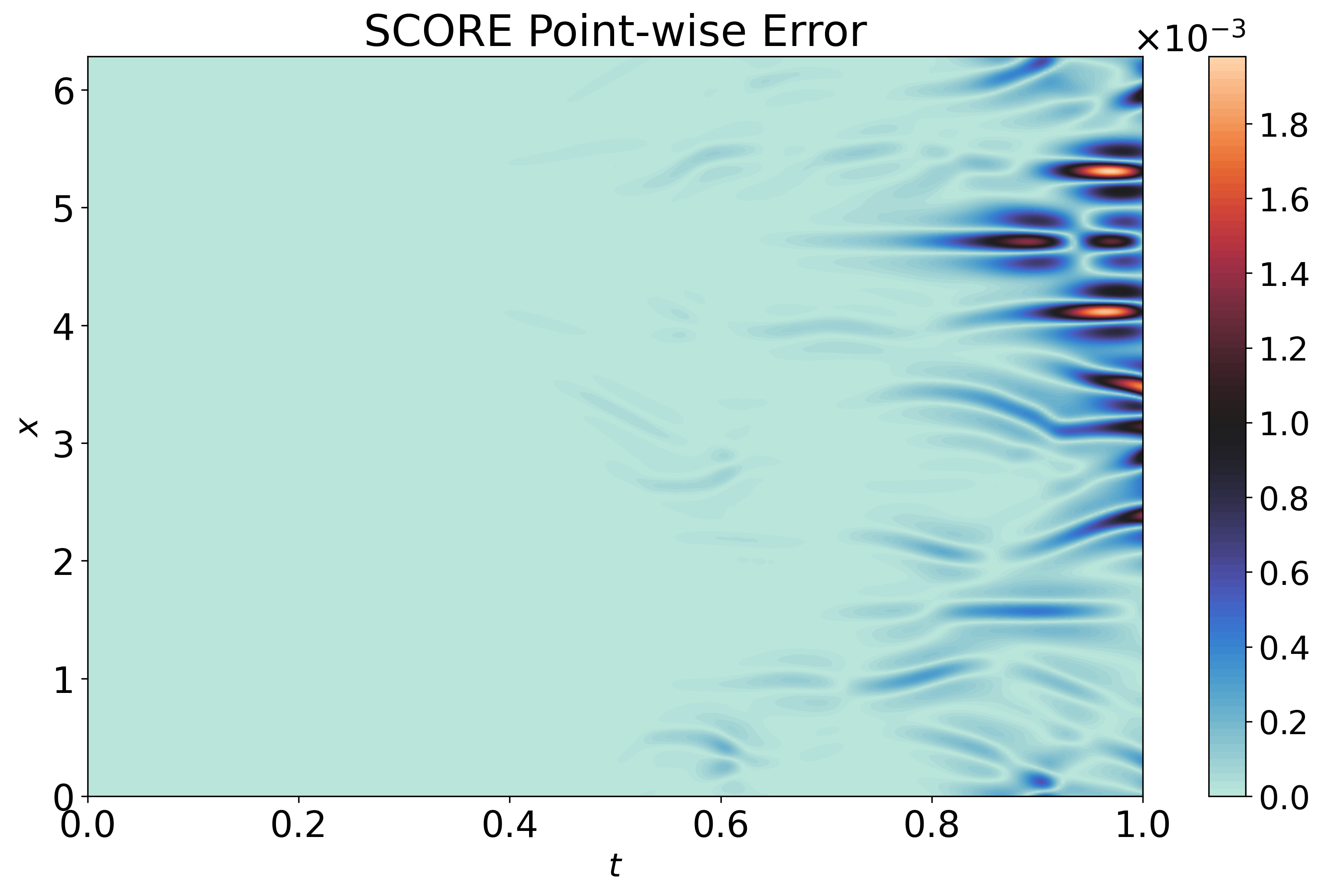} 
   \end{minipage} 

   \begin{minipage}{0.45\textwidth}
     \centering
     \includegraphics[width=\linewidth]{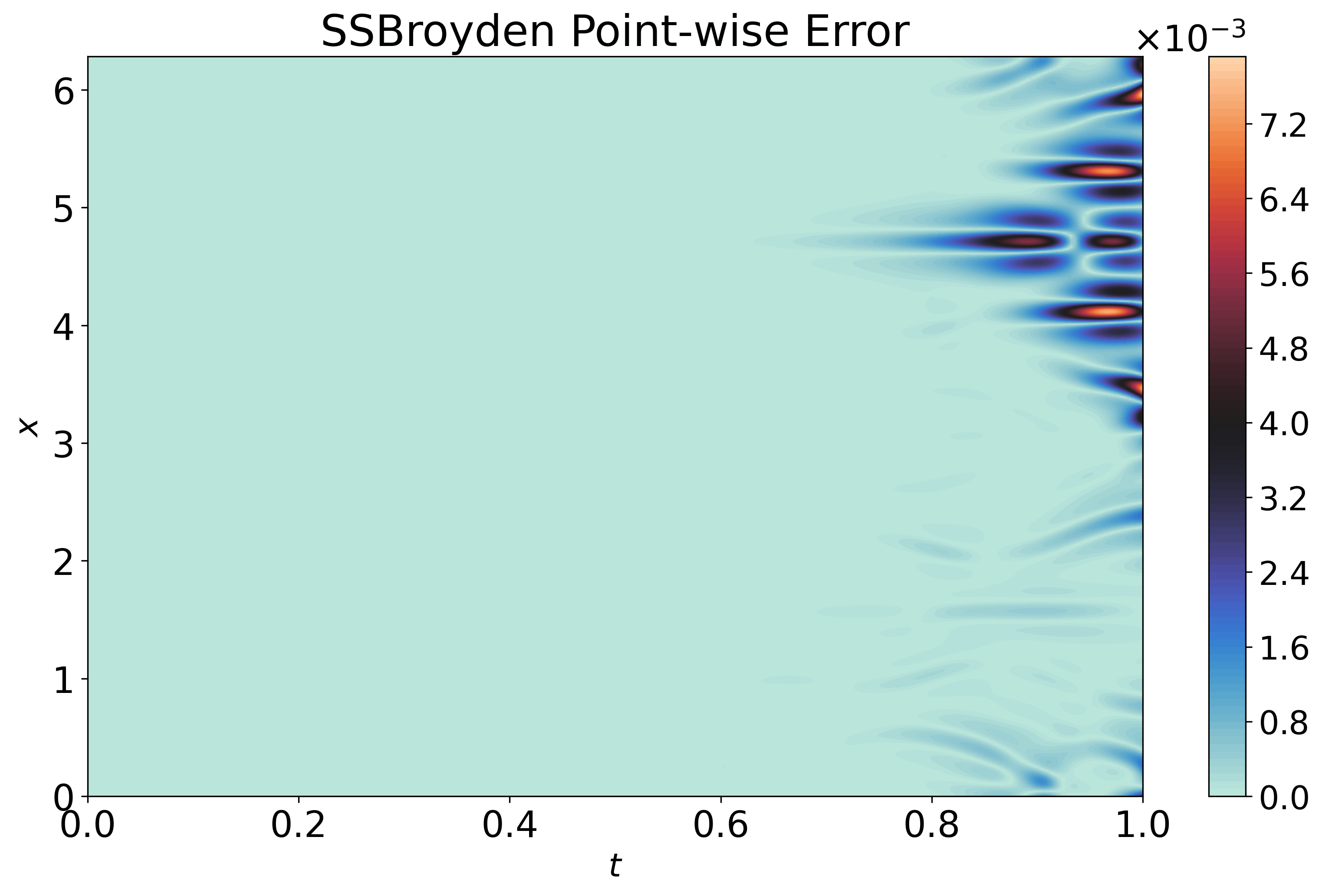} 
   \end{minipage}  
   \begin{minipage}{0.45\textwidth}
     \centering
     \includegraphics[width=\linewidth]{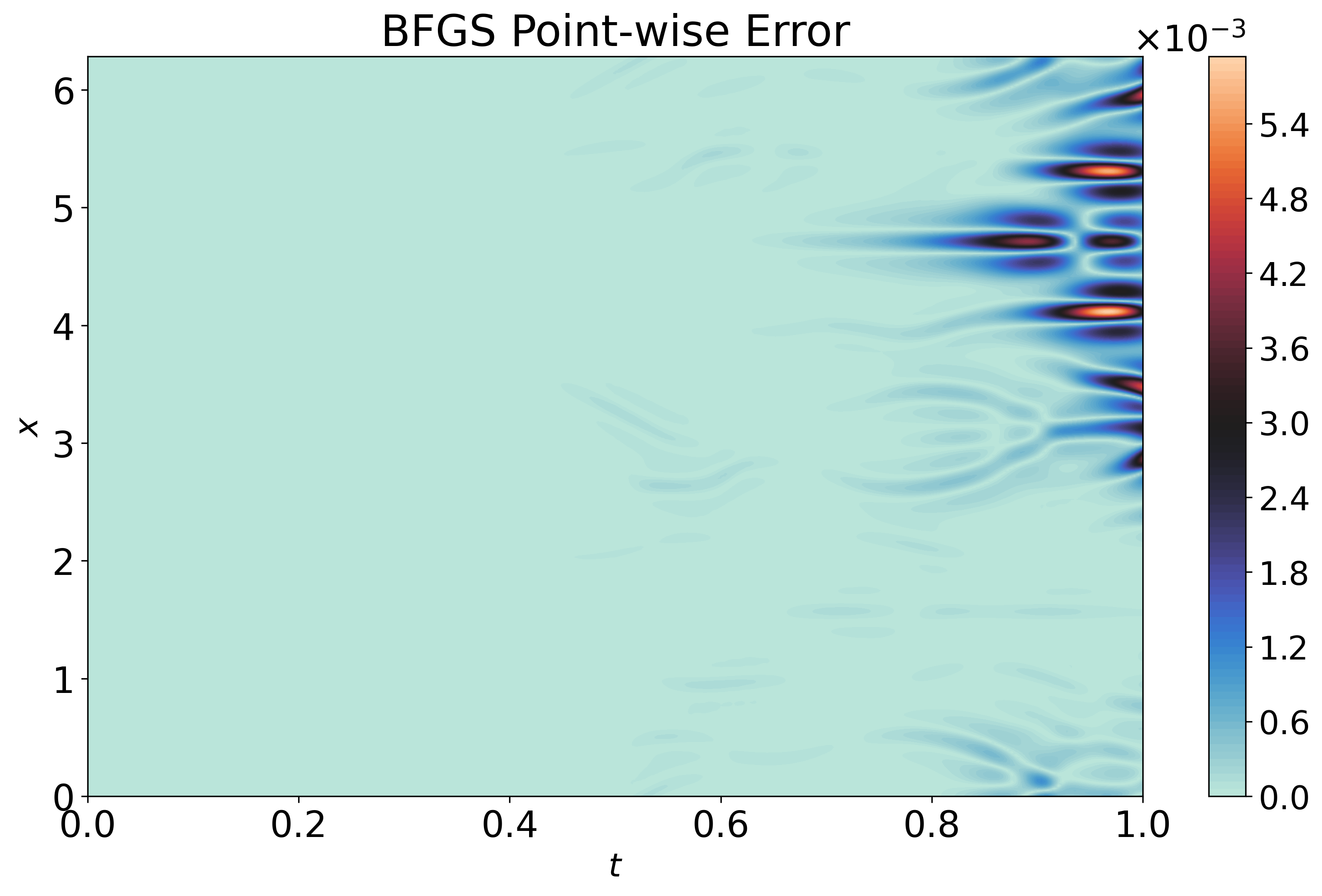} 
   \end{minipage} 
   
    \caption{{\bf Spatiotemporal heatmaps for the Kuramoto-Sivashinsky equation.} The panels show the reference solution together with the absolute error fields obtained by \textsc{SCORE}, SSBroyden, and BFGS over the full space-time domain. Since all three methods produce visually indistinguishable predictions, we report the error fields rather than the predicted solutions themselves. All error maps use the same color scale for a fair visual comparison.}\label{fig:ks_vis}
\end{figure}

\begin{figure}[!htb]
\centering
   \begin{minipage}{\textwidth}
     \centering
     \includegraphics[width=\linewidth]{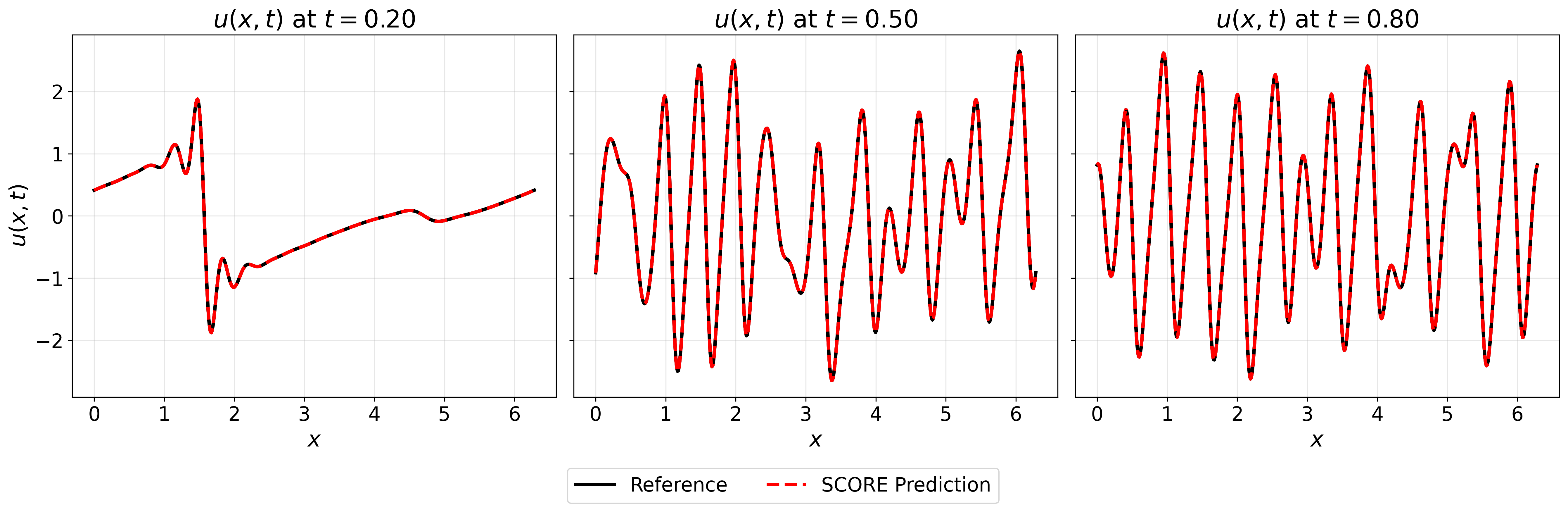} 
   \end{minipage}  
    \caption{Comparison between the reference solution and the \textsc{SCORE} prediction for the KS equation at three representative time slices, $t=0.20$, $t=0.50$, and $t=0.80$. The black solid lines represent the reference solution, and the red dashed lines represent the predicted solution. The close agreement across early, intermediate, and later times shows that \textsc{SCORE} effectively tracks the complex oscillatory spatio-temporal dynamics of the system.}

\label{fig:ks_slice}
\end{figure}

Fig.~\ref{fig:ks_vis} shows the reference solution and the corresponding point-wise absolute error fields for \textsc{SCORE}, SSBroyden, and BFGS over the full space-time domain. Since all three methods generate visually similar predictions, the error maps provide a more discriminative comparison than the predicted solutions themselves. Across the early time windows, all methods achieve very small errors. As time evolves, however, the KS dynamics develop increasingly rich oscillatory structures, and the later windows become noticeably more sensitive to approximation and propagation errors. In this regime, \textsc{SCORE} yields the most concentrated and lowest-magnitude error field, while SSBroyden and BFGS exhibit larger error regions, especially near the final windows. Moreover, this global comparison is consistent with the representative time-slice plots in Fig.~\ref{fig:ks_slice}. At $t=0.20$, $t=0.50$, and $t=0.80$, the \textsc{SCORE} prediction remains closely aligned with the reference solution, accurately capturing both the phase and amplitude of the spatial oscillations.

\begin{table}[t]
\centering
\caption{Quantitative error comparison for the Kuramoto-Sivashinsky equation. Lower values indicate better accuracy.}
\label{tab:ks_error}
\begin{tabular}{lcc}
\toprule
Optimizer & Relative $L^2$ Error & $L^\infty$ Error\\
\midrule
\textsc{SCORE}   & $\mathbf{1.84\times 10^{-4}}$ & $\mathbf{1.96\times 10^{-3}}$\\
BFGS        & $5.59\times 10^{-4}$ & $7.78\times 10^{-3}$ \\
SSBroyden             & $7.95\times 10^{-4}$ & $5.91\times 10^{-3}$\\
\bottomrule
\end{tabular}
\end{table}

\begin{figure}[!htb]
\centering
   \begin{minipage}{0.6\textwidth}
     \centering
     \includegraphics[width=\linewidth]{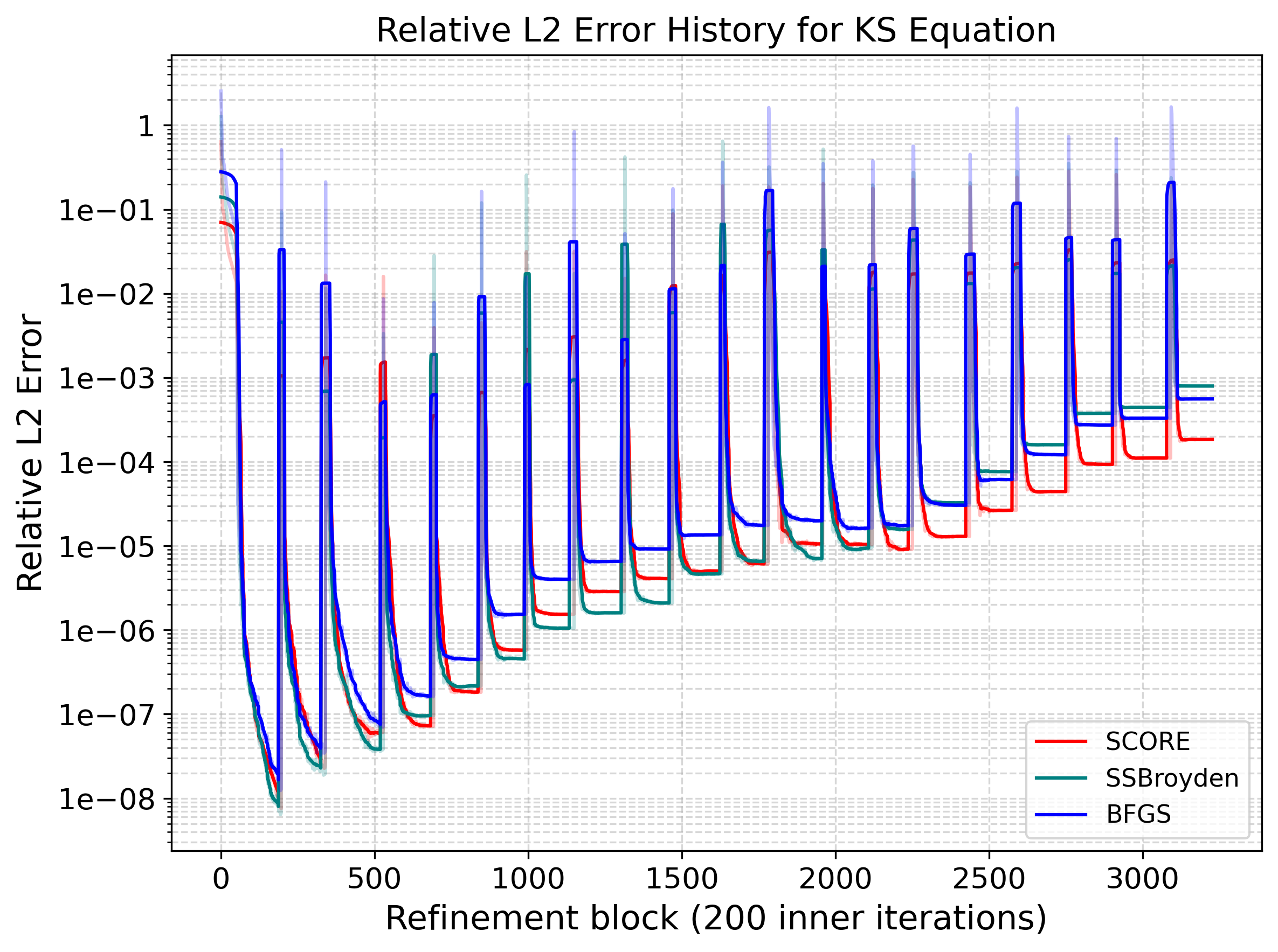} 
   \end{minipage}  
    \caption{{\bf Relative $L^2$ error history for the KS equation.}  Each sharp decrease corresponds to second-order refinement within a window, while each jump reflects the transition to the next window in the time-marching procedure. As training progresses, the attainable error floor rises and continued improvement becomes increasingly difficult. In this late-stage small-residual regime, \textsc{SCORE} consistently sustains further error reduction more effectively than conventional quasi-Newton baselines.}\label{fig:ks_hist}

\end{figure}

The optimization advantage of \textsc{SCORE} is most clearly seen in the relative $L^2$ error history shown in Fig.~\ref{fig:ks_hist}. Because training is performed window by window, the error trajectory displays a repeated drop-and-reset pattern, where each sharp decrease corresponds to refinement within a window and each jump marks the transition to the next one. While all methods reduce the error effectively in the early windows, the optimization becomes progressively more difficult in later windows, where the attainable error floor increases and additional improvement becomes harder to obtain. In this late-stage, small-residual regime, \textsc{SCORE} consistently reaches a lower terminal error than SSBroyden and BFGS. The advantage is relatively mild at the beginning but becomes increasingly pronounced as the temporal marching proceeds, indicating that \textsc{SCORE} is particularly effective at sustaining continued refinement after conventional quasi-Newton methods begin to stagnate.

This behavior is well aligned with the motivation in \S\ref{sec:shifted-secant-motivation}. The later windows of the KS problem are important not only because they contain richer oscillatory dynamics, but also because they correspond to a regime in which the residual is already small and further progress depends on stable second-order refinement. The empirical evidence therefore supports the central premise behind \textsc{SCORE}: its benefit appears most clearly when optimization is no longer limited by coarse descent, but by the ability to make reliable additional progress in a delicate low-residual regime.

\subsection{Korteweg--de Vries Equation}
\label{sec:kdv}

We next evaluate \textsc{SCORE} on the one-dimensional Korteweg--de Vries (KdV) equation, a classical nonlinear dispersive PDE that is widely used to assess the ability of PINNs to capture wave propagation and soliton-like dynamics. The problem is given by
\begin{subequations}
\begin{align}
&u_t + \eta u u_x + \mu^2 u_{xxx} = 0, \quad t \in (0,1),\; x \in (-1,1),\\
&u(x,0) = \cos(\pi x), \quad x \in (-1,1),\\
&u(t,-1) = u(t,1), \quad t \in (0,1),
\end{align}
\end{subequations}
where $u$ denotes the wave amplitude, $\eta$ controls the nonlinear interaction strength, and $\mu$ determines the dispersion level. Following the standard setting used in the literature, we fix $\eta = 1$ and $\mu = 0.022$.

For all three optimizers, we use the same PINN architecture and the same first-stage training setup to ensure a fair comparison. The network is a fully connected MLP with three hidden layers of width 80, $\tanh$ activations, and a scalar output. To encode spatial periodicity, we use the same periodic Fourier feature embedding as in the Burgers experiment, with domain length $L_x = 2$ and embedding degree $m=5$. This construction makes the network output periodic in space by design.

All methods are warm-started using Adam for 5{,}000 epochs with an initial learning rate of $10^{-3}$ and an exponential decay schedule with decay rate 0.9 over 5{,}000 steps. During the quasi-Newton refinement phase, we use $750$ outer iterations, each allowing up to $200$ inner quasi-Newton steps, yielding a nominal budget of $150{,}000$ refinement iterations. At each outer iteration, 30{,}000 collocation points are sampled uniformly from the spatio-temporal domain.

\begin{figure}[!htb]
\centering
   \begin{minipage}{0.45\textwidth}
     \centering
     \includegraphics[width=\linewidth]{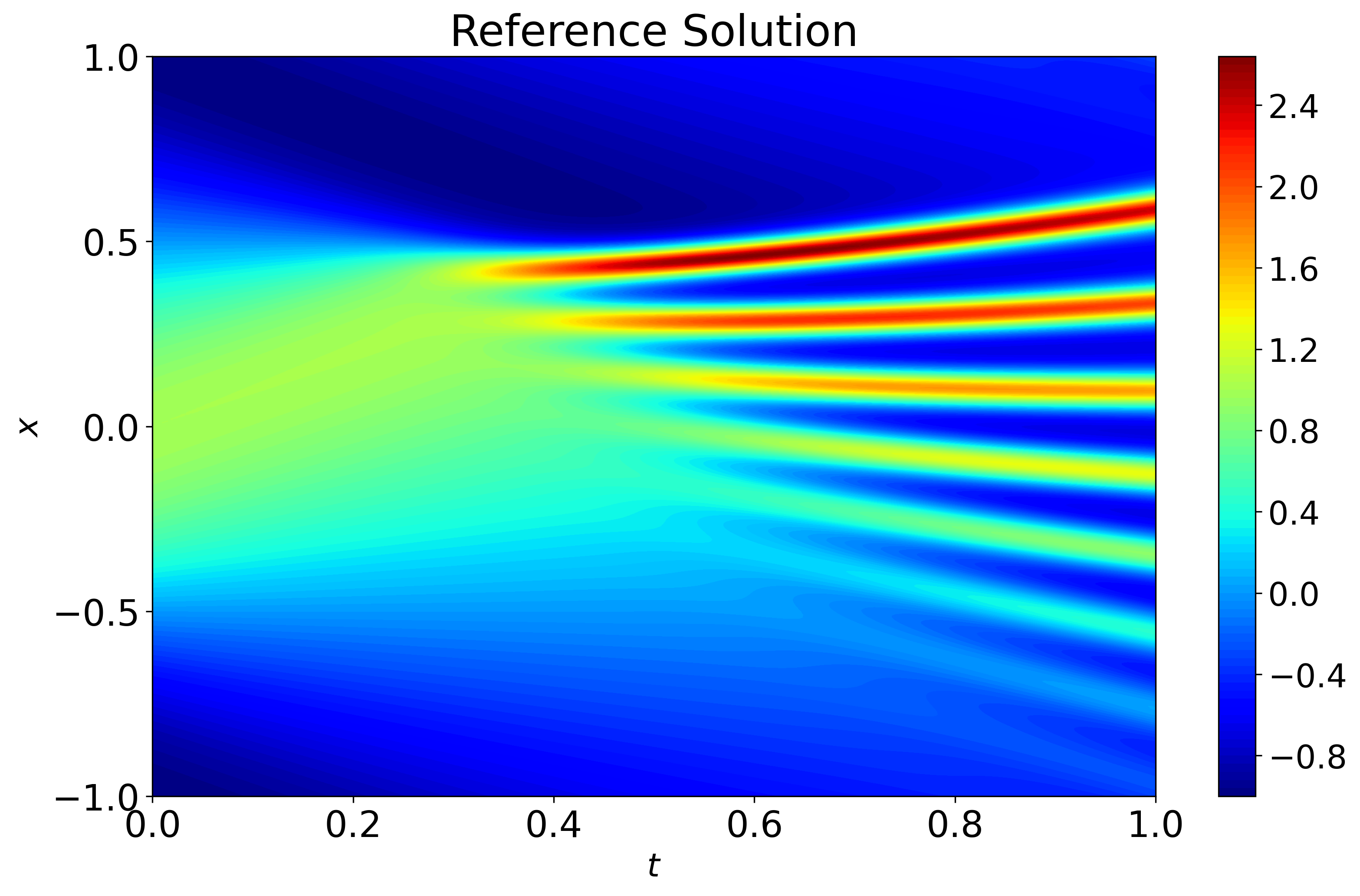} 
   \end{minipage}  
   \begin{minipage}{0.45\textwidth}
     \centering
     \includegraphics[width=\linewidth]{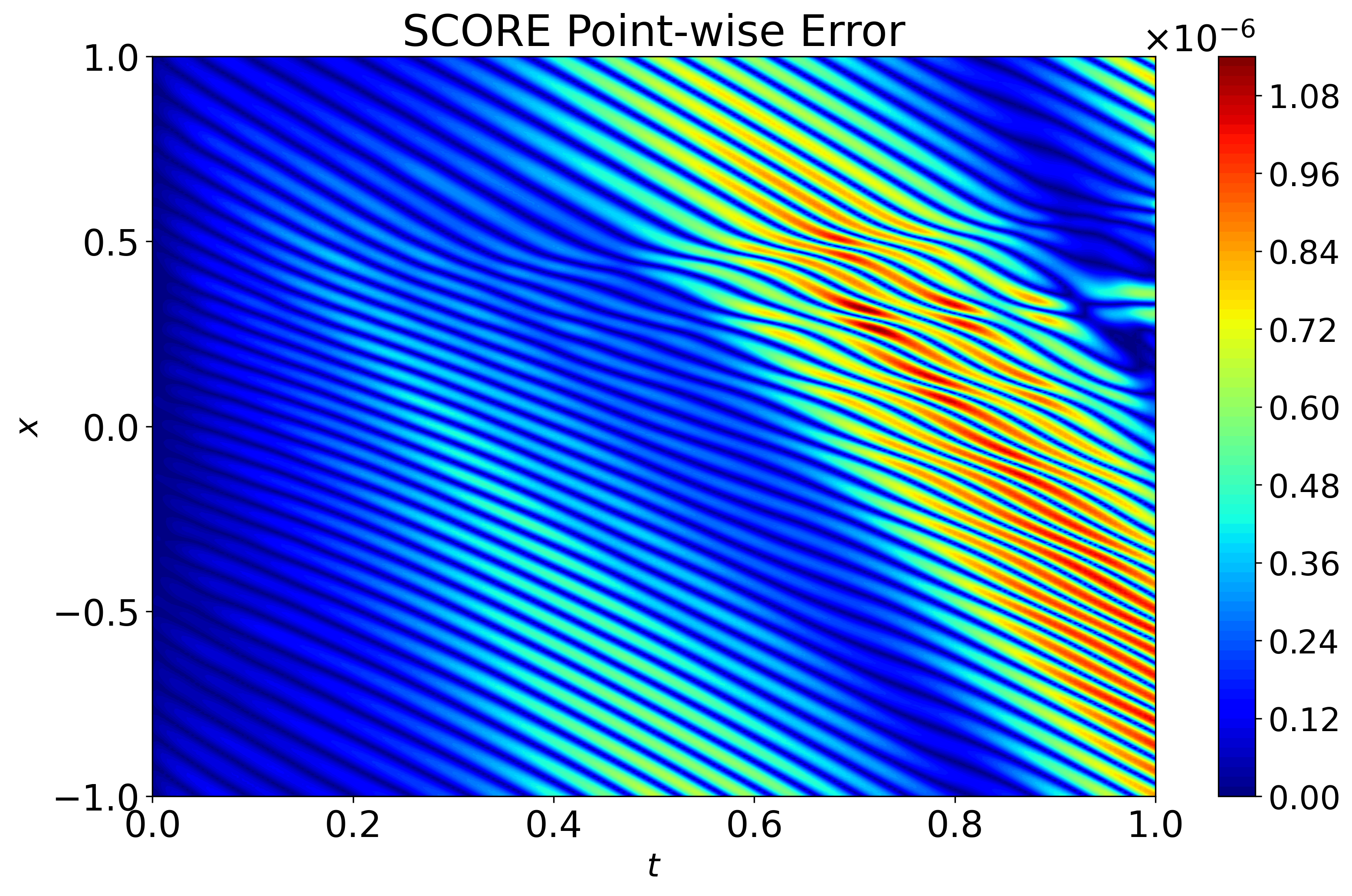} 
   \end{minipage} 

   \begin{minipage}{0.45\textwidth}
     \centering
     \includegraphics[width=\linewidth]{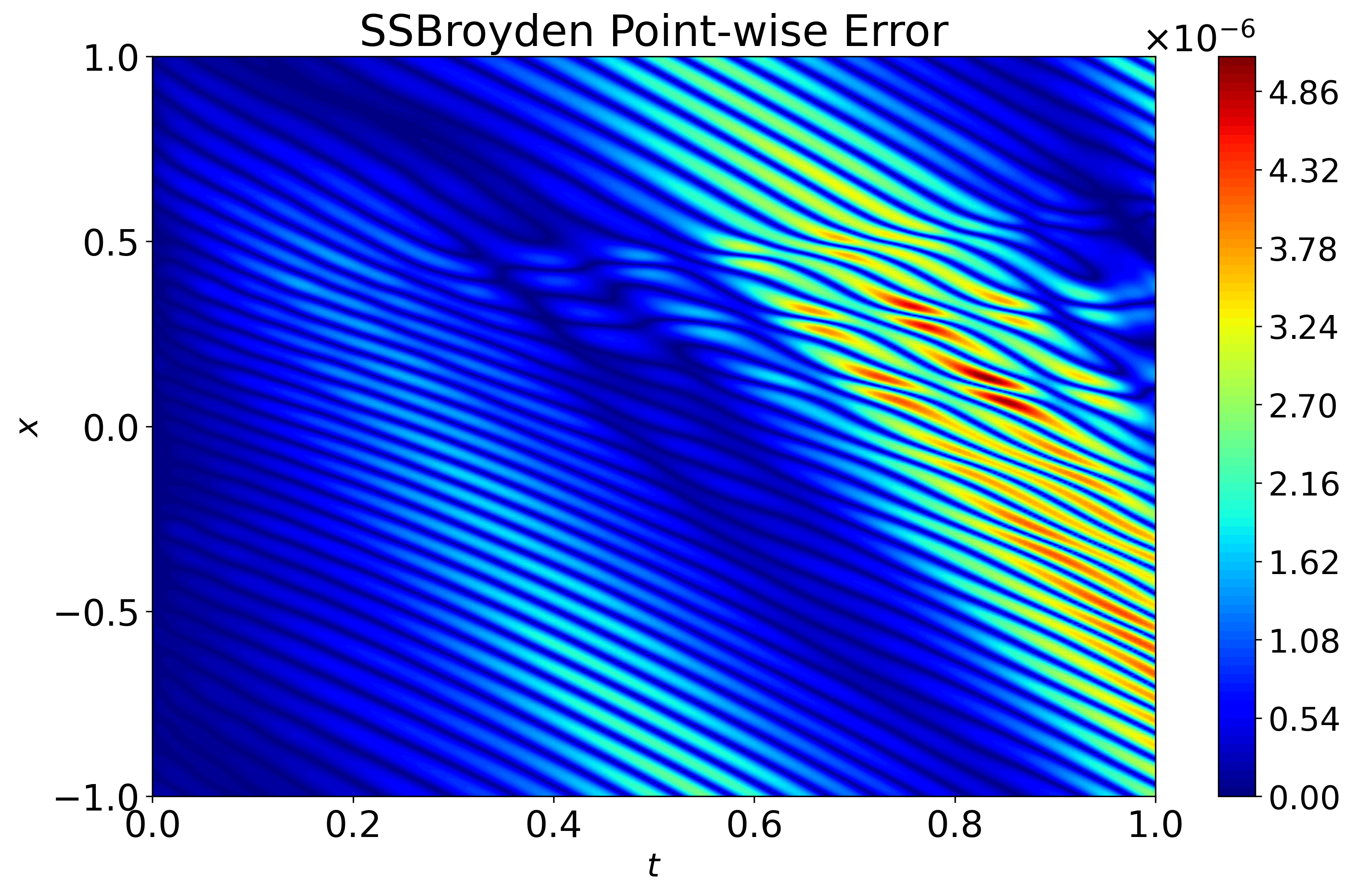} 
   \end{minipage}  
   \begin{minipage}{0.45\textwidth}
     \centering
     \includegraphics[width=\linewidth]{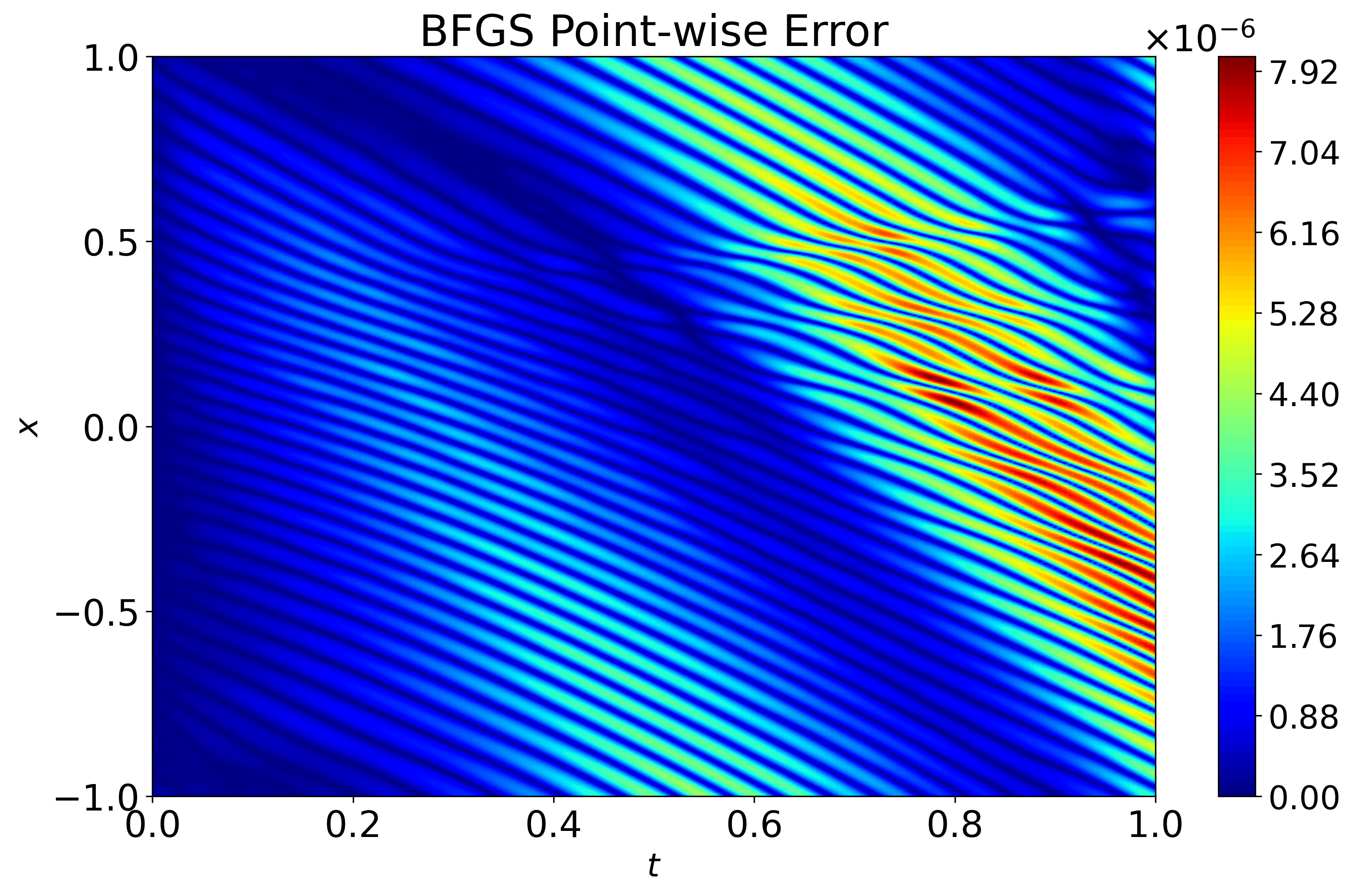} 
   \end{minipage} 
   
    \caption{{\bf Spatiotemporal heatmaps for the Korteweg-de Vries equation.} The panels show the reference solution together with the absolute error fields obtained by \textsc{SCORE}, SSBroyden, and BFGS over the full space-time domain. Since all three methods produce visually indistinguishable predictions, we report the error fields rather than the predicted solutions themselves. All error maps use the same color scale for a fair visual comparison.}\label{fig:kdv_vis}
\end{figure}

The results for the KdV equation consistently demonstrate the advantage of \textsc{SCORE} over the competing quasi-Newton optimizers. As shown in Fig.~\ref{fig:kdv_vis}, although all three methods produce visually accurate predictions, the error heatmaps reveal that \textsc{SCORE} achieves a cleaner and less intense error distribution over the full spatiotemporal domain. Its residual errors are especially better controlled in regions with more complex wave interactions and sharper solution variations, whereas SSBroyden and BFGS exhibit more visible structured error patterns. This suggests that \textsc{SCORE} provides more accurate local refinement and better captures the fine-scale dynamics of the KdV solution.

\begin{figure}[!htb]
\centering
   \begin{minipage}{0.6\textwidth}
     \centering
     \includegraphics[width=\linewidth]{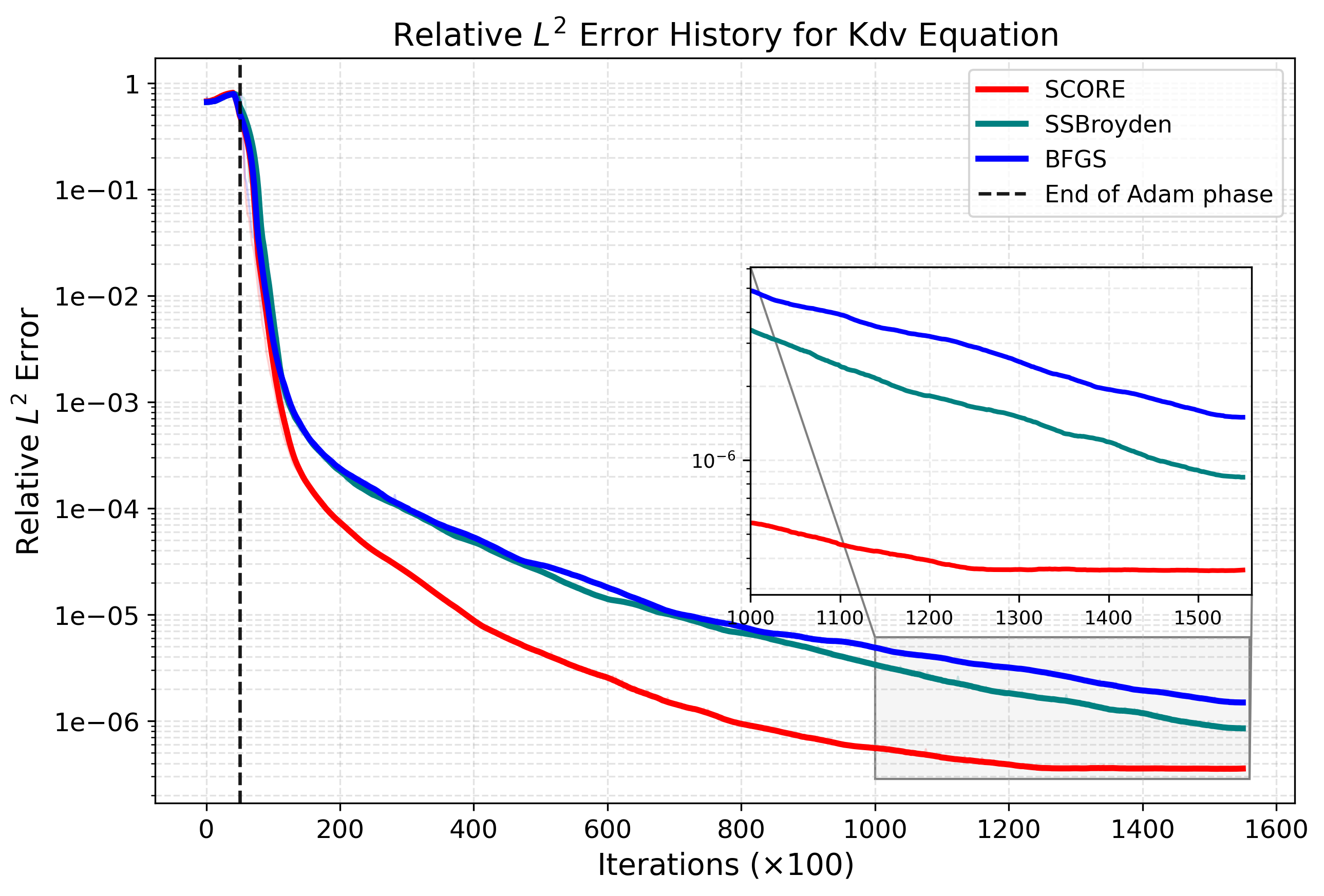} 
   \end{minipage}  
    \caption{{\bf Relative $L^2$ error history for the KdV equation.} The dashed vertical line marks the end of the Adam warm-start phase and the beginning of quasi-Newton refinement.}
\label{fig:kdv_hist}
\end{figure}

This observation is further supported by the convergence behavior in Fig.~\ref{fig:kdv_hist} and the quantitative comparison in Table~\ref{tab:kdv_error}. After the shared Adam warm-start phase, \textsc{SCORE} enters a more favorable refinement trajectory, decreases the relative $L^2$ error more effectively, and reaches a lower final error floor than the competing methods. Table~\ref{tab:kdv_error} confirms that \textsc{SCORE} delivers the best overall accuracy among the tested optimizers while maintaining essentially the same computational cost. Overall, these results show that \textsc{SCORE} achieves a better accuracy--efficiency trade-off and is particularly effective for PINN training on nonlinear dispersive PDEs such as the KdV equation.

As with the Burgers case, this behavior is in line with the motivation in \S~\ref{sec:shifted-secant-motivation}, but it stresses a complementary aspect of the method. The KdV equation combines nonlinear advection with third-order dispersion, which induces strong directional anisotropy in the loss landscape during second-order refinement. In this regime, the secant displacement carries informative but heterogeneous curvature information across directions, and a fixed worst-case calibration tends to either under-use or over-shrink it. By coupling the shift magnitude to the quasi-Newton decrement, \textsc{SCORE} adapts to the realized curvature scale at each step, which we believe is the main reason for the systematic improvement observed during the late refinement stage on this dispersive problem.

\begin{table}[t]
\centering
\caption{Quantitative error comparison for the KdV equation. Lower values indicate better accuracy.}
\label{tab:kdv_error}
\begin{tabular}{lcc}
\toprule
Optimizer & Relative $L^2$ Error & $L^\infty$ Error \\ 
\midrule
\textsc{SCORE}   & $\mathbf{3.57\times 10^{-7}}$ & $\mathbf{1.21\times 10^{-6}}$\\
BFGS        & $1.49\times 10^{-6}$ & $8.05\times 10^{-6}$ \\
SSBroyden             & $8.57\times 10^{-7}$ & $5.13\times 10^{-6}$\\
\bottomrule
\end{tabular}
\end{table}

\subsection{Complex Ginzburg--Landau Equation}
\label{sec:cgl}

We next evaluate \textsc{SCORE} on the two-dimensional complex Ginzburg--Landau equation, a canonical nonlinear PDE for modeling dissipative pattern formation, oscillatory instabilities, and complex spatio-temporal dynamics. Owing to its coupled nonlinear and diffusive structure, this problem provides a challenging benchmark for PINNs in higher-dimensional settings. In complex form, the equation is given by
$$
\frac{\partial A}{\partial t}
=
\varepsilon \Delta A + kA - k(1+1.5i)A|A|^2,
\qquad
t \in (0,1), \quad (x,y) \in (-1,1)^2,
$$
subject to periodic boundary conditions, where $A=A(x,y,t)$ denotes the complex-valued field, $\varepsilon$ is the diffusion coefficient, and $k$ controls the linear growth and nonlinear saturation strength.

In our experiments, we set $\varepsilon = 0.1$ and $k = 20$. The initial condition is chosen as
$$
A(x,y,0)
=
(10y + 10ix)\exp\left(-0.01\left(2500x^2 + 2500y^2\right)\right).
$$
To obtain the reference solution, we generate high-resolution simulation data on the spatial domain $[-1,1]^2$ over the time interval $[0,1]$ using a Fourier spectral solver with periodic boundary conditions. Writing $A=u+iv$, we use the resulting real and imaginary components $u$ and $v$ as the target fields for evaluation.

For this problem, we represent the complex field as $A=u+iv$ and train the PINN to predict the two real-valued components $u(x,y,t)$ and $v(x,y,t)$ simultaneously. The network is a fully connected MLP with three hidden layers of width 50 and a two-dimensional output. As in \S~\ref{sec:ks}, we normalize the temporal input within each training window and use periodic Fourier feature embeddings to encode the spatial coordinates. In particular, for each of $x$ and $y$ we use a Fourier feature embedding of degree $m=2$, and concatenate the resulting spatial features with the normalized time variable before feeding them into the MLP.

We follow the same time-marching training strategy as in \S~\ref{sec:ks}, but use a coarser temporal decomposition with $5$ consecutive windows over $[0,1]$. Within each window, we sample $30{,}000$ collocation points in the interior of the space-time subdomain and impose the initial condition on the full spatial grid with penalty weight $\lambda_{\mathrm{IC}} = 500$. The optimization schedule is also kept the same as in the KS experiment: Adam warm-up is used only in the first window, followed by second-order refinement in every window for $30{,}000$ inner iterations with the stochastic batch refreshed every $200$ steps, and each new window is initialized from the learned parameters of the previous one.

\begin{figure}[!htb]
\centering

%==================== Row 1: Prediction ====================
\begin{minipage}{0.23\textwidth}
  \centering
  \includegraphics[width=\linewidth]{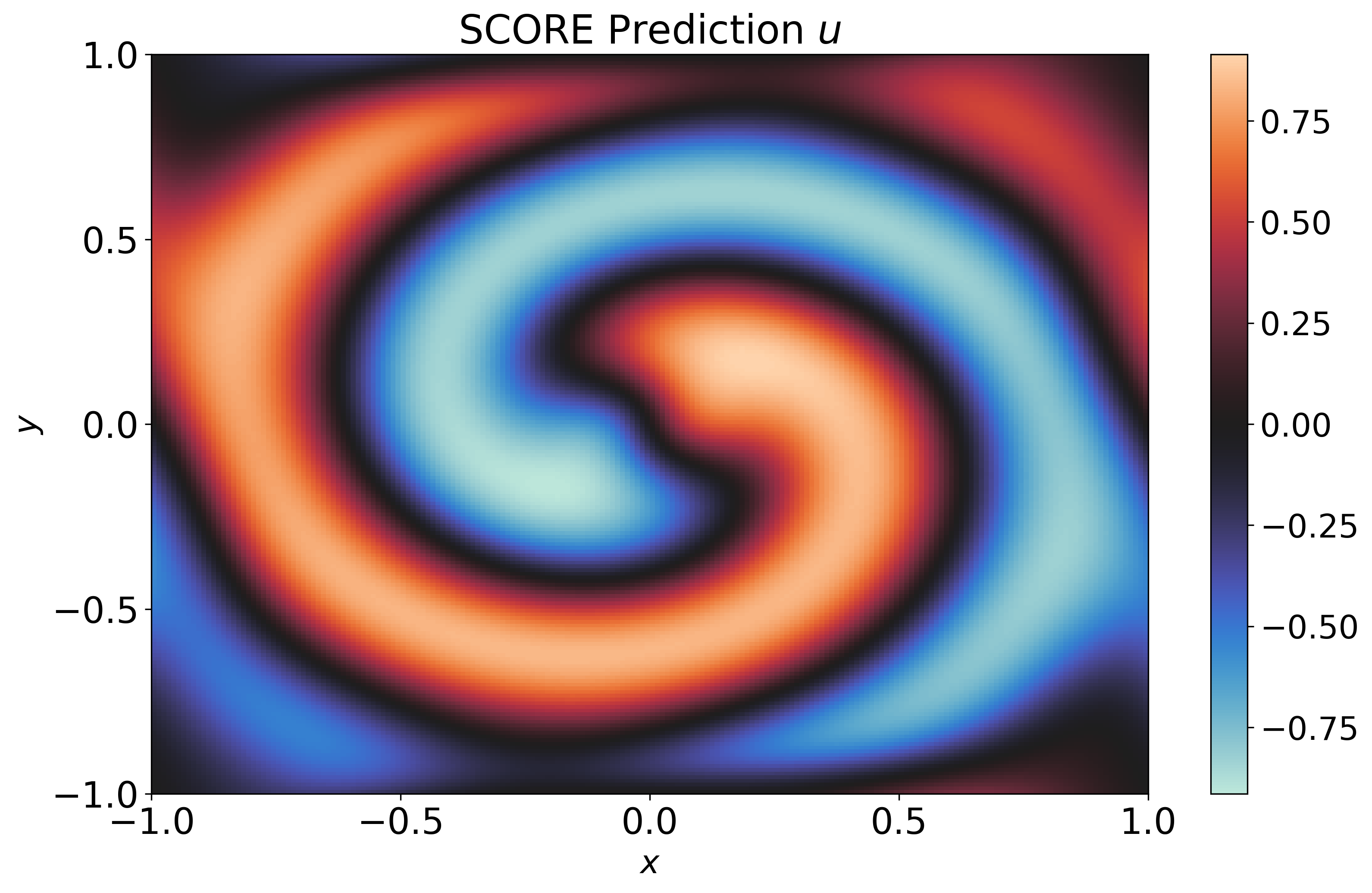}
\end{minipage}
\hfill
\begin{minipage}{0.23\textwidth}
  \centering
  \includegraphics[width=\linewidth]{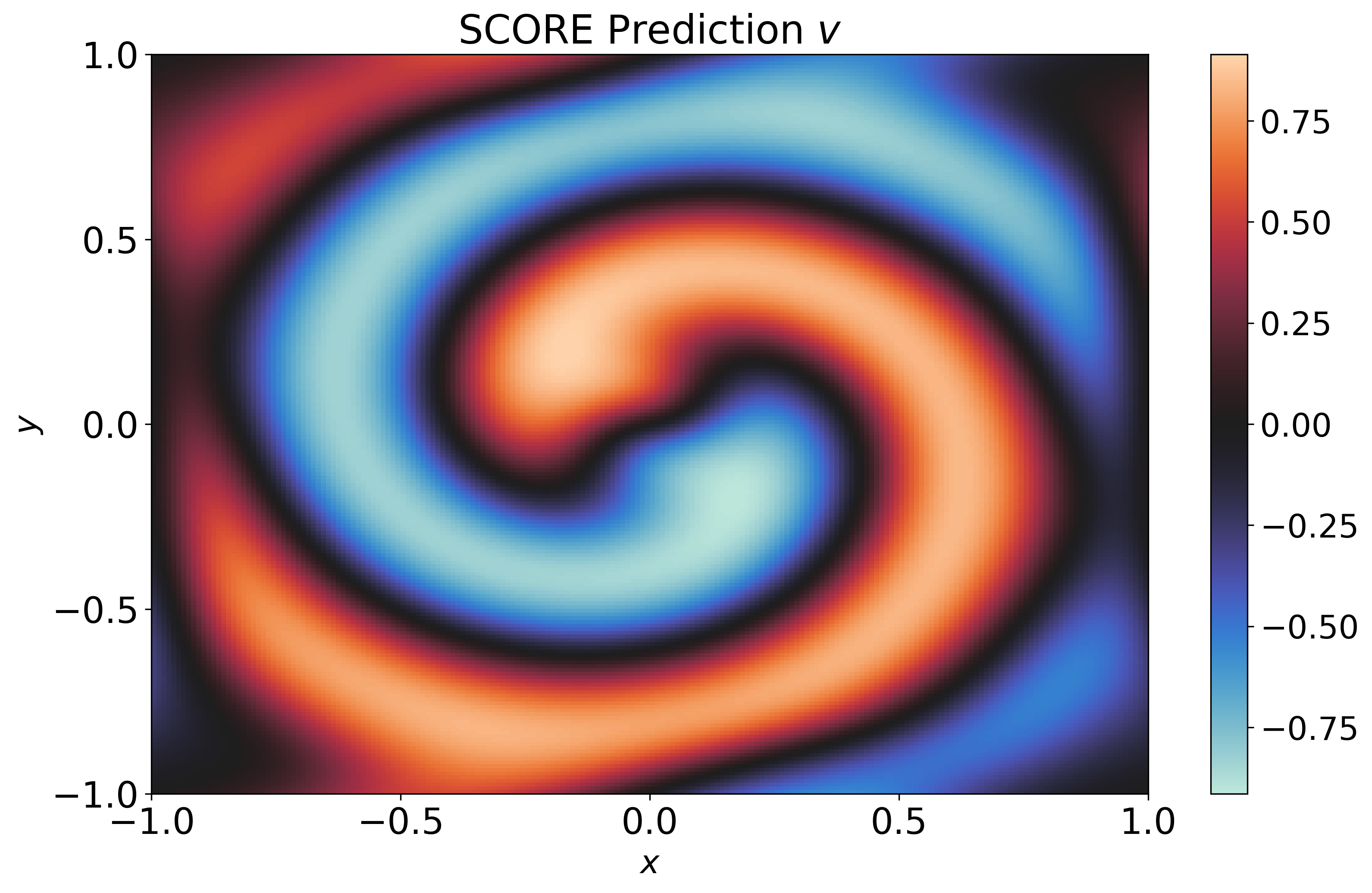}
\end{minipage}
\hfill
\begin{minipage}{0.23\textwidth}
  \centering
  \includegraphics[width=\linewidth]{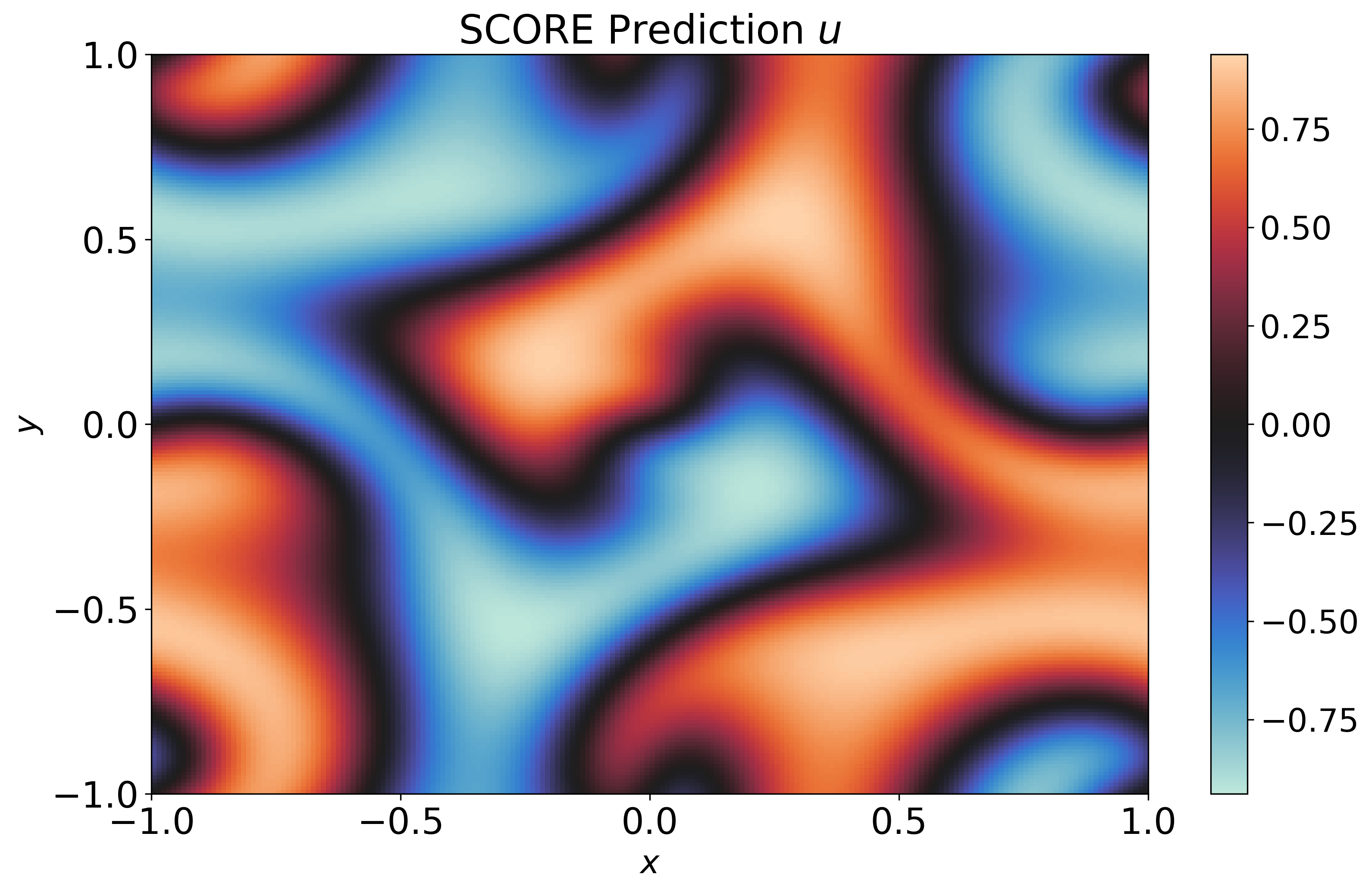}
\end{minipage}
\hfill
\begin{minipage}{0.23\textwidth}
  \centering
  \includegraphics[width=\linewidth]{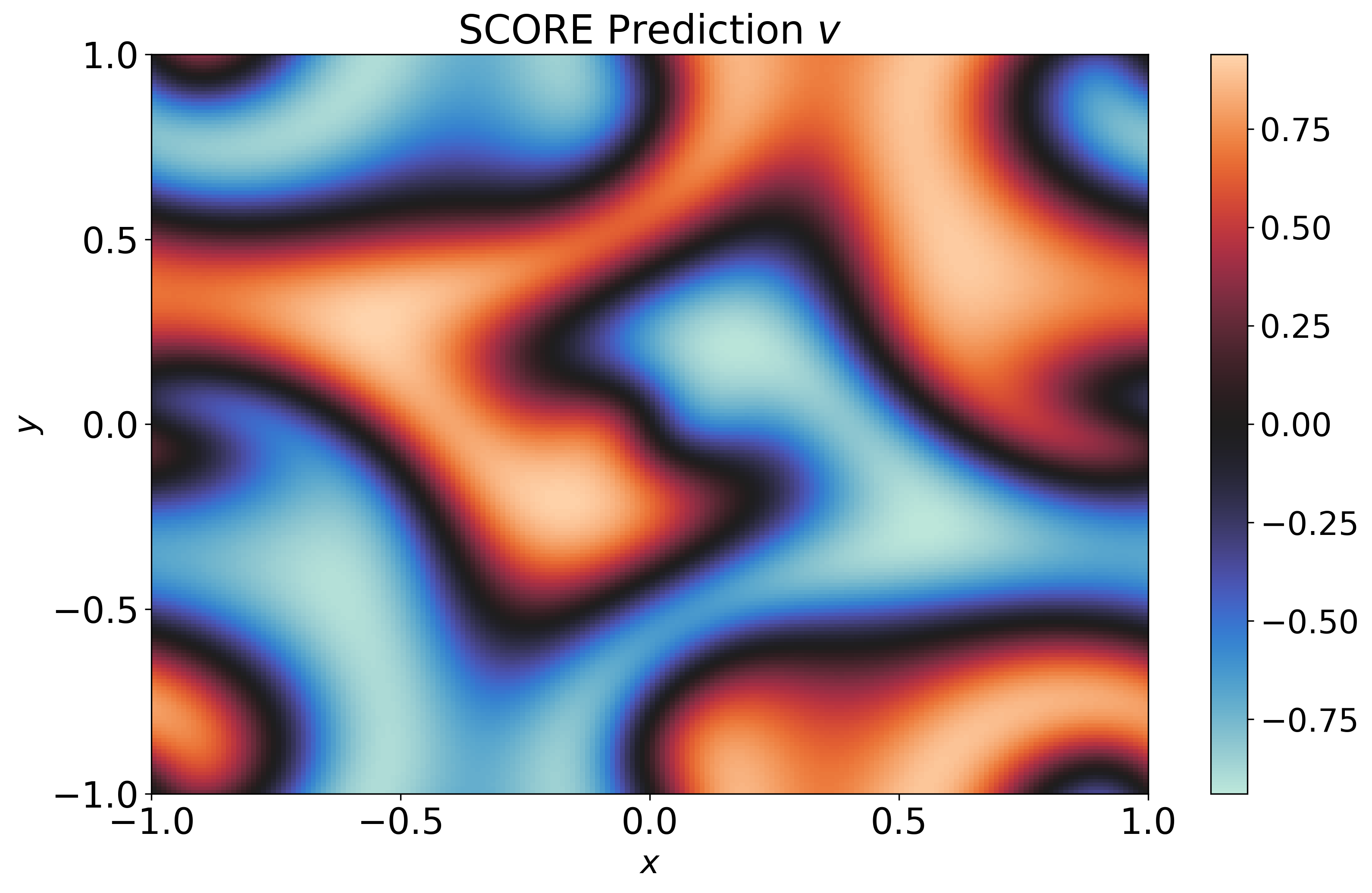}
\end{minipage}

\vspace{0.5em}

%==================== Row 2: Error ====================
\begin{minipage}{0.23\textwidth}
  \centering
  \includegraphics[width=\linewidth]{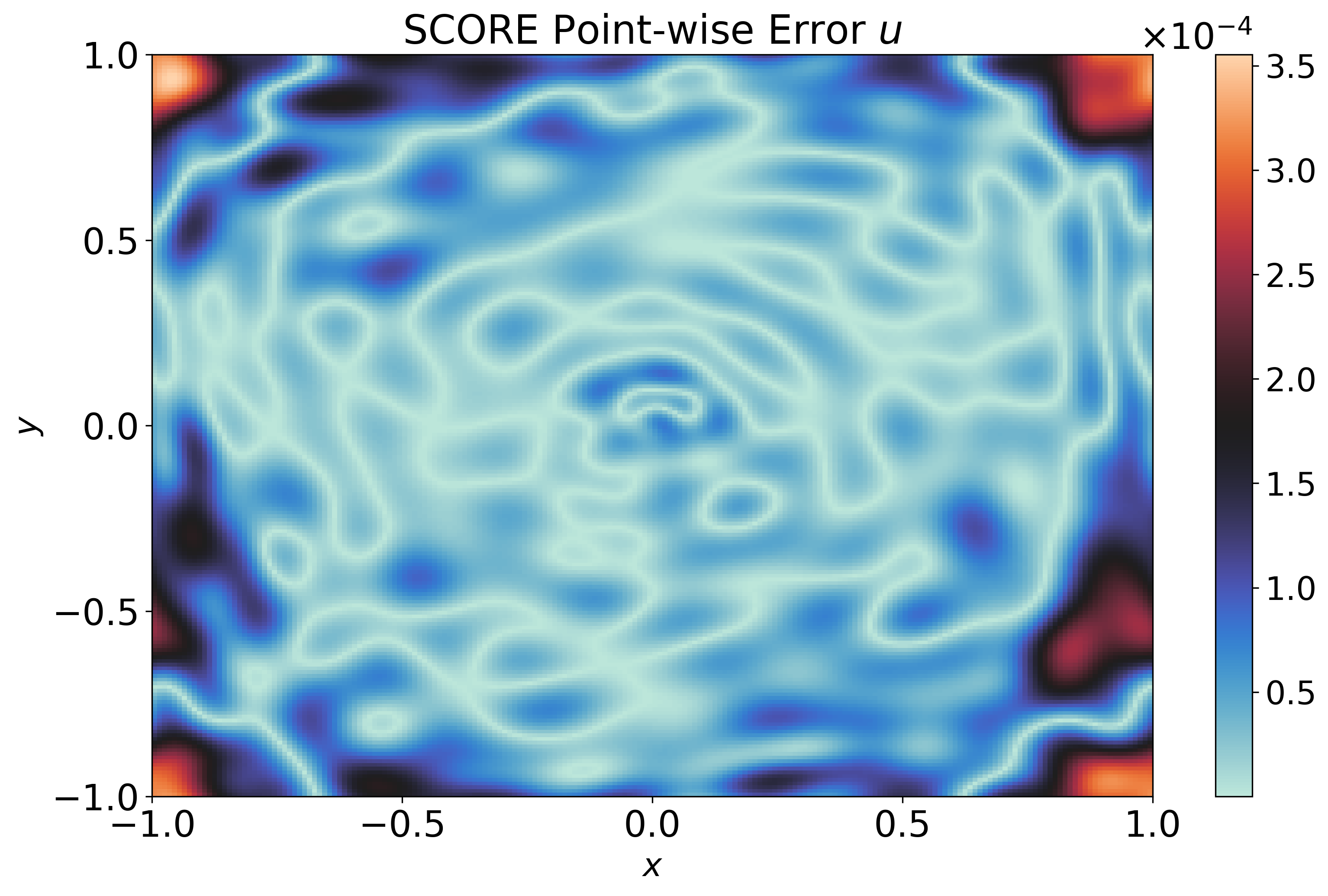}
\end{minipage}
\hfill
\begin{minipage}{0.23\textwidth}
  \centering
  \includegraphics[width=\linewidth]{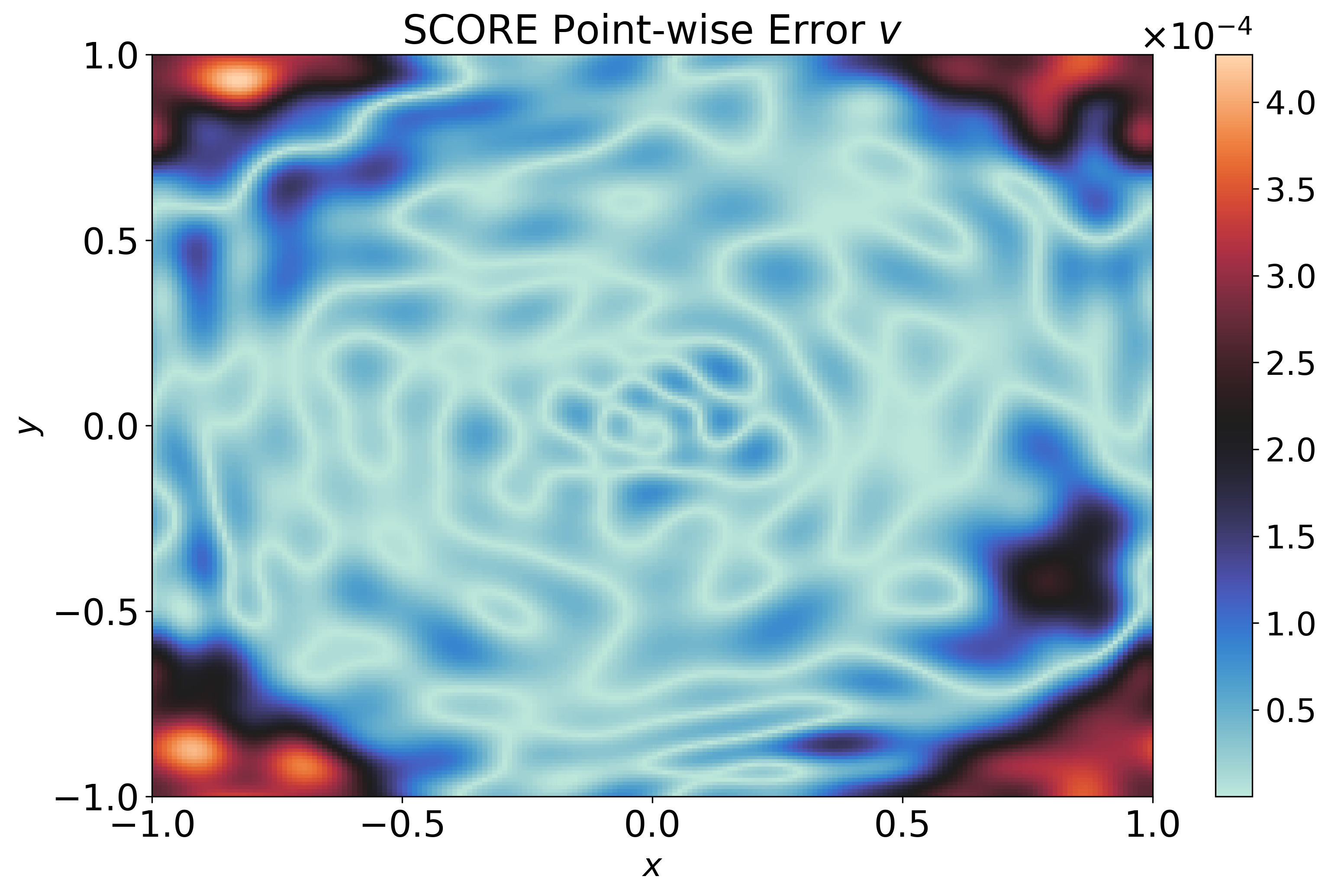}
\end{minipage}
\hfill
\begin{minipage}{0.23\textwidth}
  \centering
  \includegraphics[width=\linewidth]{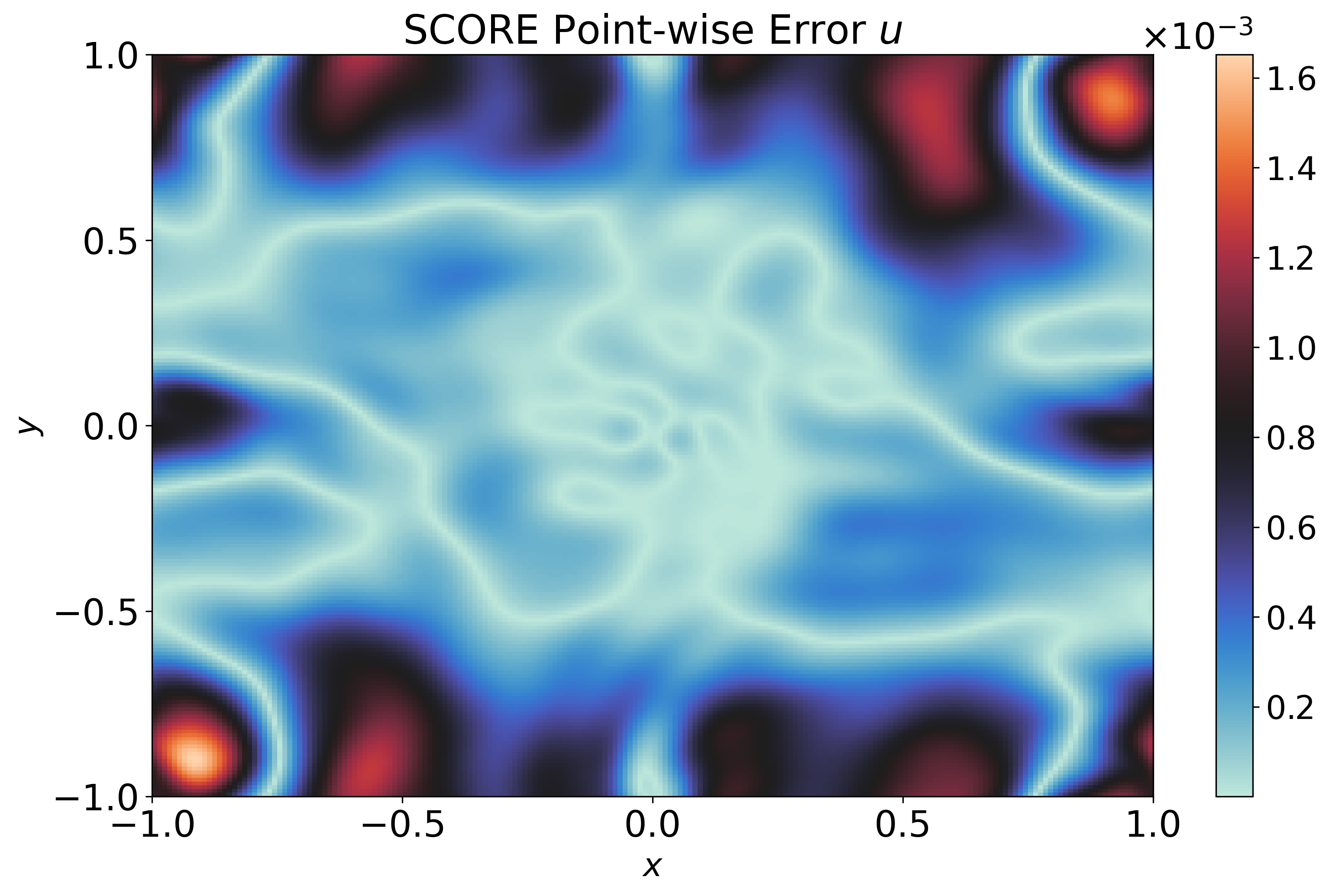}
\end{minipage}
\hfill
\begin{minipage}{0.23\textwidth}
  \centering
  \includegraphics[width=\linewidth]{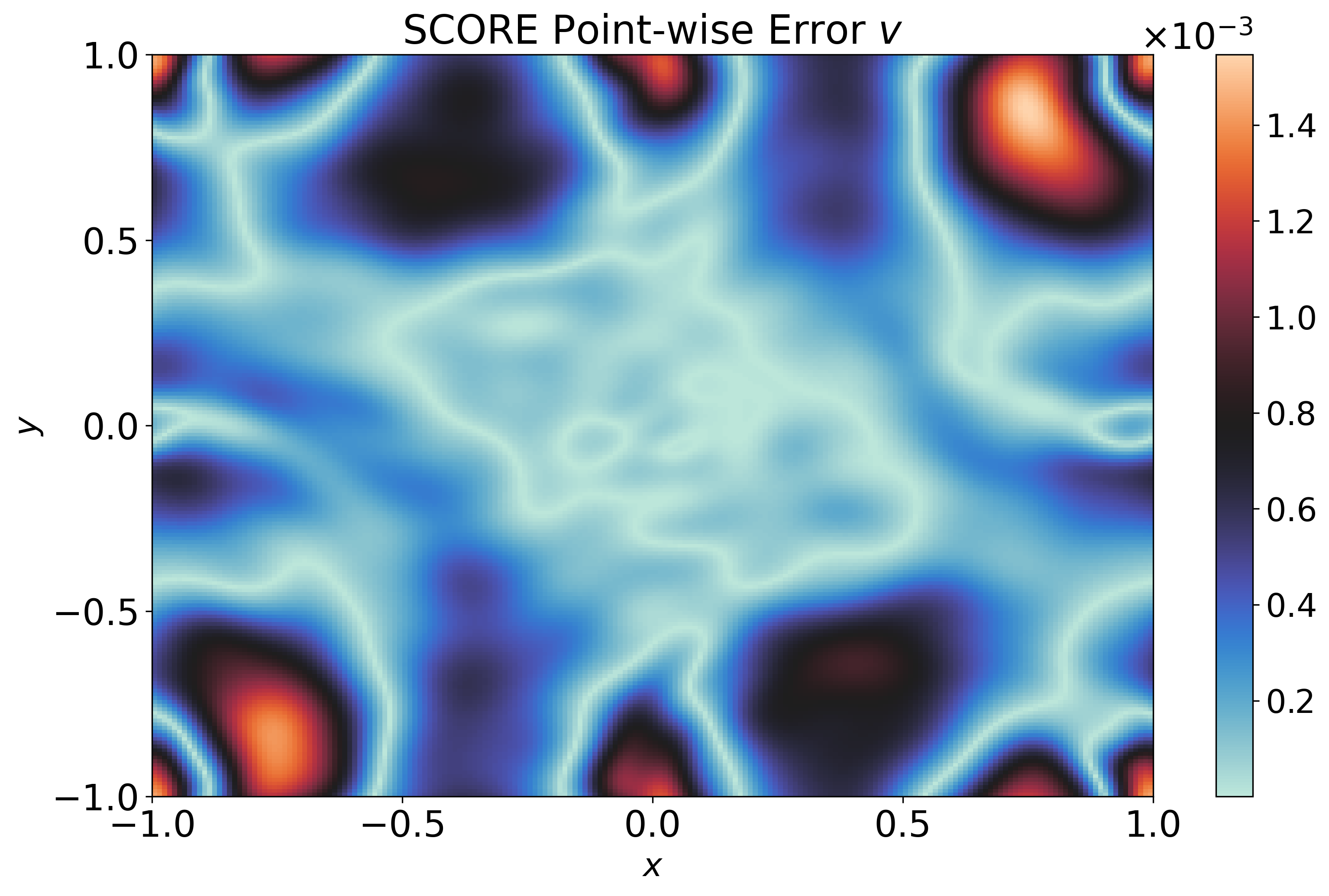}
\end{minipage}

\caption{Heatmaps of the predicted solution and point-wise absolute error for the two-dimensional complex Ginzburg--Landau equation at two representative time instants, $t=0.4$ and $t=0.8$. The first row shows the \textsc{SCORE} predictions, while the second row reports the corresponding absolute errors. From left to right, the columns correspond to the real component $u$ at $t=0.4$, the imaginary component $v$ at $t=0.4$, the real component $u$ at $t=0.8$, and the imaginary component $v$ at $t=0.8$.}
\label{fig:cgl_vis}
\end{figure}

The results on the complex Ginzburg--Landau equation further demonstrate the advantage of \textsc{SCORE} in a more challenging two-dimensional and complex-valued setting. As shown in Fig.~\ref{fig:cgl_vis}, \textsc{SCORE} accurately captures both the real and imaginary components of the solution at the representative time instants. The predicted patterns are well aligned with the reference dynamics, while the corresponding absolute error maps remain uniformly small and localized, indicating that the method preserves the main spatio-temporal structures of the evolving complex field with high fidelity. This is particularly meaningful for the GL equation, where nonlinear interaction, diffusion, and oscillatory behavior are tightly coupled and can easily amplify approximation errors over time.

\begin{figure}[!htb]
\centering
   \begin{minipage}{0.6\textwidth}
     \centering
     \includegraphics[width=\linewidth]{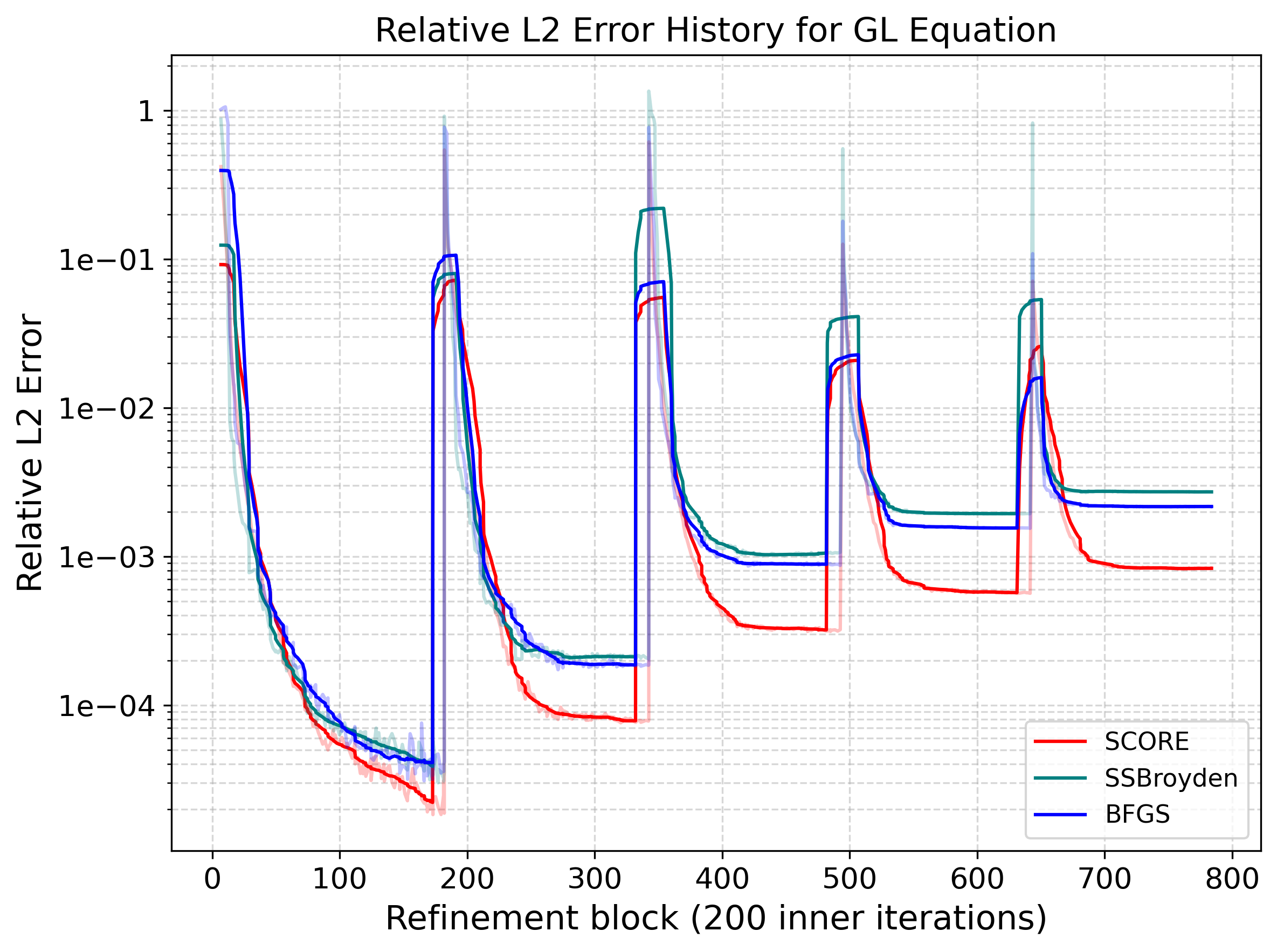} 
   \end{minipage}  
    \caption{{\bf Relative $L^2$ error history for the complex GL equation.} Since the solution is complex-valued and represented by its real and imaginary components, $u$ and $v$, we report a single relative $L^2$ error curve computed from the combined error of both fields rather than showing separate histories for each component. This provides a unified measure of the approximation quality of the complex solution throughout training.}
\label{fig:cgl_hist}
\end{figure}

The convergence and quantitative results further support this observation. In Fig.~\ref{fig:cgl_hist}, we report a single relative $L^2$ error curve computed from the complex field as a whole. Writing $A=u+iv$ and $\hat A=\hat u+i\hat v$, the plotted error is defined as
\begin{align*}
E_{\mathrm{CGL}}
=
\frac{
\sqrt{
\sum_{k=1}^{N}
\left(
|\hat u(\mathbf{x}_k,t_k)-u(\mathbf{x}_k,t_k)|^2
+
|\hat v(\mathbf{x}_k,t_k)-v(\mathbf{x}_k,t_k)|^2
\right)
}
}{
\sqrt{
\sum_{k=1}^{N}
\left(
|u(\mathbf{x}_k,t_k)|^2
+
|v(\mathbf{x}_k,t_k)|^2
\right)
}
}.
\end{align*}
This aggregated metric jointly measures the discrepancy in the two real-valued components $u$ and $v$, whereas Table~\ref{tab:gl_error} reports the componentwise errors for $u$ and $v$ separately.
 This aggregated metric is more appropriate here because the PINN predicts the two components simultaneously and they together represent one complex solution field. Under this evaluation, \textsc{SCORE} shows a more favorable convergence trajectory during training. This trend is consistent with Table~\ref{tab:gl_error}, which shows that \textsc{SCORE} achieves the best overall accuracy on both components and under both error metrics among the compared optimizers. Overall, these results indicate that \textsc{SCORE} remains highly effective in higher-dimensional complex-valued PDEs and provides a clear improvement in solution quality over SSBroyden and BFGS.

\begin{table}[t]
\centering
\caption{Quantitative error comparison for the complex GL equation. Lower values indicate better accuracy.}
\label{tab:gl_error}
\begin{tabular}{lcccc}
\toprule
Optimizer & Relative $L^2$ Error ($u$) & Relative $L^2$ Error ($v$) & $L^\infty$ Error ($u$) & $L^\infty$ Error ($v$) \\
\midrule
\textsc{SCORE} & $\mathbf{8.42\times 10^{-4}}$ & $\mathbf{8.46\times 10^{-4}}$ & $\mathbf{1.98\times 10^{-3}}$ & $\mathbf{1.65\times 10^{-3}}$ \\
BFGS          & $2.19\times 10^{-3}$ & $2.14\times 10^{-3}$ & $7.05\times 10^{-3}$ & $8.23\times 10^{-3}$ \\
SSBroyden     & $2.71\times 10^{-3}$ & $2.73\times 10^{-3}$ & $9.13\times 10^{-3}$ & $8.66\times 10^{-3}$ \\
\bottomrule
\end{tabular}
\end{table}

\section{Conclusion}
\label{sec:conclusion}

This work introduced \textsc{SCORE}, a self-concordant quasi-Newton optimizer for improving the late-stage refinement of physics-informed neural networks. The method is motivated by the observation that PINN training often enters a small-residual but highly ill-conditioned regime, where conventional Lipschitz-calibrated step-size rules may become overly conservative. Instead of relying on worst-case curvature variation, \textsc{SCORE} uses a curvature-relative perspective inspired by weak self-concordance to adapt the optimizer to the local geometry encountered along the training trajectory.

The key idea behind \textsc{SCORE} is to realize a locally shifted metric directly through the secant displacement used in the quasi-Newton update. This shifted secant geometry provides a simple way to stabilize curvature information without changing the objective or introducing an external penalty term. By coupling the shift magnitude to the quasi-Newton decrement, the method adjusts its curvature stabilization according to the current optimization state. The same decrement also defines a self-concordant candidate step, which is tested before falling back to the standard Wolfe line search. In this way, \textsc{SCORE} links step selection and metric stabilization through a single local geometric quantity.

Experiments on the viscous Burgers, Kuramoto--Sivashinsky, Korteweg--de Vries, and complex Ginzburg--Landau equations demonstrate that \textsc{SCORE} consistently improves high-accuracy PINN refinement compared with standard quasi-Newton baselines. Across these benchmarks, the method attains lower prediction errors while maintaining comparable block-level runtime under the same outer--inner training protocol. These results support the central premise of the paper: in the late stages of PINN optimization, further progress is often limited not by coarse descent, but by the ability to extract reliable curvature information from a delicate local loss geometry.

More broadly, this work points to a curvature-relative view of PINN optimization. While \textsc{SCORE} instantiates this idea through a shifted secant geometry, the same principle may support a broader class of second-order methods designed specifically for physics-informed objectives. Future work may explore how weak self-concordance, residual decomposition, adaptive collocation, and blockwise curvature diagnostics can be combined to produce optimization algorithms whose step sizes and metric updates are governed by the geometry of the PDE loss itself. We believe this direction offers a promising route toward robust high-accuracy solvers for scientific machine learning.

\section*{CRediT authorship contribution statement}
\textbf{Chenhao Si:} Conceptualization, Methodology, Investigation, Software, Writing - original draft;
\textbf{Kang An:} Methodology, Writing - original draft;
\textbf{Shiqian Ma:} Conceptualization, Methodology, Supervision, Writing – review and editing.
\textbf{Ming Yan:} Conceptualization, Methodology, Supervision, Writing - review and editing;

\section*{Declaration of competing interest}
The authors declare that they have no known competing financial interests or personal relationships that could have appeared to influence the work reported in this paper.

\section*{Data availability}
The codes generated during the current study will be available upon reasonable request after the article is published.
\section*{Acknowledgments}
Chenhao Si and Ming Yan were partially supported by the National Natural Science Foundation of China (72495131, 82441027), Guangdong Provincial Key Laboratory of Mathematical Foundations for Artificial Intelligence (2023B1212010001), Shenzhen Stability Science Program, and the Shenzhen Science and Technology Program (JCYJ20250604141043020).

\section*{Declaration of generative AI use}
During the preparation of this work the author(s) used ChatGPT 5.6 Pro in order to polish the sentences only. After using this tool/service, the author(s) reviewed and edited the content as needed and take(s) full responsibility for the content of the published article.

\bibliographystyle{unsrt}  
\bibliography{ref}

\end{document}